\documentclass{article}

\usepackage{microtype}
\usepackage{graphicx}
\usepackage{subfigure}
\usepackage{booktabs} 

\usepackage{graphicx}
\usepackage{amsmath}
\usepackage{amsthm}
\usepackage{amssymb}
\usepackage{bm}
\usepackage[abs]{overpic}
\usepackage{algorithm}
\usepackage[noend]{algorithmic}
\usepackage{eqparbox}
\renewcommand\algorithmiccomment[1]{%
  \hfill\#\ \eqparbox{COMMENT}{#1}%
}
\renewcommand\algorithmicthen{}
\renewcommand\algorithmicdo{}
\usepackage{bbold}
\usepackage{enumitem}
\usepackage{xcolor}

\newtheorem{theorem}{Theorem}
\newtheorem{lemma}{Lemma}

\newcommand{\charfct}{\mathbb{1}}

\newcommand{\R}{\mathbb{R}}
\newcommand{\N}{\mathbb{N}}

\DeclareMathOperator*{\argmin}{argmin} 
\DeclareMathOperator*{\argmax}{argmax}
\DeclareMathOperator*{\minimize}{minimize}
\newcommand{\fair}{\text{fair}}

\definecolor{mygreen}{rgb}{0,0.6,0}

\usepackage{makecell}

\usepackage{hyperref}

\usepackage[accepted]{my_arx_style}

\icmltitlerunning{Fair $k$-Center Clustering for Data Summarization}

\begin{document}

\twocolumn[
\icmltitle{Fair $k$-Center Clustering for Data Summarization}



\icmlsetsymbol{equal}{*}

\begin{icmlauthorlist}
\icmlauthor{Matth\"{a}us Kleindessner}{ru}
\icmlauthor{Pranjal Awasthi}{ru}
\icmlauthor{Jamie Morgenstern}{gt}
\end{icmlauthorlist}


\icmlaffiliation{ru}{Department of Computer Science, Rutgers University, NJ}
\icmlaffiliation{gt}{College of Computing, Georgia Tech, GA}

\icmlcorrespondingauthor{Matth\"{a}us Kleindessner}{matthaeus.kleindessner@rutgers.edu}
\icmlcorrespondingauthor{Pranjal Awasthi}{pranjal.awasthi@rutgers.edu}
\icmlcorrespondingauthor{Jamie Morgenstern}{jamiemmt@cs.gatech.edu}



\icmlkeywords{Machine Learning, Clustering, k-Center, Fairness, ICML}

\vskip 0.3in
]



\printAffiliationsAndNotice{}  

\begin{abstract}
In data summarization we want to choose $k$ prototypes 
in order to summarize 
a
  data set. 
We study a setting where 
the  
data set 
comprises 
several demographic groups and 
we 
are restricted to 
choose $k_{i}$ prototypes 
belonging to 
group~$i$. 
A common approach to 
the problem without the fairness constraint is to optimize a centroid-based clustering objective such as $k$-center.
A natural extension then is to incorporate 
the fairness constraint into the clustering 
problem. 
Existing algorithms for doing so run in time 
super-quadratic in the size of the data set, which is in contrast to the standard $k$-center 
problem being approximable in linear time. In this paper, we resolve this gap by providing a simple 
approximation 
 algorithm for 
 the $k$-center problem 
 under 
 the 
 fairness constraint with running time 
 linear in 
 the size of the data set and $k$. 
 If the number of demographic groups is small, the approximation guarantee of our algorithm only incurs a constant-factor overhead. 
\end{abstract}

\section{Introduction}\label{section_introduction}

Machine learning (ML) algorithms have been rapidly adopted in numerous human-centric domains, from personalized advertising to lending to health care. Fast on the heels of this ubiquity have come a whole host of concerning behaviors 
from 
these algorithms: facial recognition has higher accuracy on white, male faces \citep{Buolamwini2018}; 
online advertisements 
suggesting  arrest  are shown more frequently to search queries that  
comprise a name primarily associated with minority groups 
\citep{Sweeney2013}; 
and criminal recidivism tools 
are likely to mislabel black low-risk defendants as high-risk while mislabeling white high-risk defendants as low-risk \citep{Angwin2018}. 
There are also several examples of unsavory ML behavior pertaining  to unsupervised learning tasks, 
such as gender stereotypes  in word2vec embeddings \citep{bolukbasi2016man}. 
Most of the  academic work on fairness in ML, however, has investigated how to solve classification tasks subject to various constraints on the 
behavior of a classifier on different demo\-graphic groups~\citep[e.g., ][]{hardt2016,zafar2017}.

This paper  adds to the literature on fair methods 
 for unsupervised learning tasks 
  (see Section~\ref{section_related_work} for related work). 
We consider the problem of data summarization \citep{Hesabi2015} through the lens of algorithmic fairness. The goal of data summarization is to output  a small but representative subset of a 
data set. 
Think of 
an 
image database and a user entering a query that is matched by many images. Rather than presenting  
the user with all matching images, 
we only want to show a 
summary. 
In such an example, a data summary can be quite unfair on a demographic group. Indeed, 
Google Images has been found to answer the query ``CEO'' with a much higher fraction of images of  men 
compared to 
 the 
real-world  
 fraction of male CEOs \citep{kay2015}. 
 
 One approach to the problem of data summarization is  provided by centroid-based clustering, 
 such as $k$-center (formally defined in Section \ref{section_prob_statement}) or $k$-medoid 
(\citealp[][Section 14.3.10]{hastie2009}; sometimes referred to as $k$-median).
For a centroid-based clustering objective, 
an optimal clustering of a data set $S$ can be defined by $k$ points $c_1^*,\ldots,c_k^*\in S$, called centroids, such that the clusters are formed by assigning  every $s\in S$ to its closest centroid. Since the centroids are good representatives of their clusters,  
 the set of centroids can be used as a summary of $S$. This approach of data  summarization via centroid-based clustering is used
%
in numerous domains, for example in text summarization \citep{moens1999} or robotics \citep{Girdhar2012}.

 If the data set~$S$ comprises several demographic groups $S_1,\ldots,S_m$, we may consider $c_1^*,\ldots,c_k^*$ to be  a fair summary only if the groups are  represented fairly: 
 if in the real world  70\% of CEOs are male and we want to 
 output 
 ten images for the query ``CEO'', then  three of the ten images should show women. 
 Formally, this can be encoded with one parameter $k_{S_i}$ for every 
 group~$S_i$. Our goal is then to minimize the 
 clustering 
 objective under the constraint that  $k_{S_i}$ many centroids belong to $S_i$.  A constraint of this form can also enforce balanced summaries: even if in the real world there are more male CEOs than female ones, we might want to output an equal number of male and female images  
 to reflect that gender is not definitional to the~role~of~CEO.
 
 Centroid-based clustering under such a  constraint has been studied in the theoretical computer science literature (see 
Sections~\ref{section_prob_statement}~and~\ref{section_related_work}). 
 However, existing 
approximation  
 algorithms for this problem run in time $\omega(|S|^2)$, while the unconstrained $k$-center clustering problem can be approximated  in time  linear in $|S|$. Since data summarization is particularly useful for massive data sets, such a slowdown may be practically prohibitive. 
The contribution of this paper is to  present  
a simple 
approximation algorithm for $k$-center clustering under our fairness constraint with running time only linear in 
 $|S|$ and $k$. The improved running time comes at the price of a worse guarantee on the approximation factor if the number of demographic groups is large. 
However, note that in practical situations concerning fairness, the number of groups is often quite small (e.g., when the groups encode gender or race). Furthermore, in our extensive numerical simulations we \emph{never} observed a large approximation factor, even when the number of groups was large 
(cf. Section~\ref{section_experiments}), indicating the practical 
usefulness of our 
algorithm. 

\vspace{1pt}
\textbf{Outline of the paper~~} 
In Section~\ref{section_prob_statement}, we formally state the $k$-center and the fair $k$-center problem. In Section~\ref{section_algorithm}, we present our algorithm and provide a sketch of its analysis. The full proofs can be found in Appendix~\ref{appendix_proofs}. We discuss related work in Section~\ref{section_related_work} and present 
a number of 
experiments in Section~\ref{section_experiments}. Further experiments can be found in Appendix~\ref{appendix_experiments}. We conclude with a discussion in Section~\ref{section_discussion}.

\vspace{1pt}
\textbf{Notation~~} 
For $l\in\N$, we sometimes use 
$[l]=\{1,\ldots,l\}$.

\section{
Definition of 
$k$-Center and Fair  $k$-Center}\label{section_prob_statement}

Let $S$ be a finite data set and $d:S\times S\rightarrow \R_{\geq 0}$ be a metric on $S$. In particular, we assume $d$ to satisfy the triangle inequality. The 
standard 
 $k$-center clustering problem
is the   minimization problem
\begin{align}\label{unfair_k_center}
\minimize_{C=\{c_1,\ldots,c_k\}\subseteq S}~ \max_{s\in S} \;d(s,C),
\end{align}
where $k\in \N$ is a given parameter and $d(s,C)=\min_{c\in C} d(s,c)$. Here, $c_1,\ldots,c_k$ are called centers. Any set of centers defines a clustering of $S$ by assigning every $s\in S$ to its closest center. The $k$-center problem is NP-hard and is also NP-hard to approximate to a factor better than $2$ 
(\citealp{gonzalez1985}; \citealp{vazirani_approx}, Chapter~5).
The 
famous 
greedy strategy 
of 
\citet{gonzalez1985}
 is 
a $2$-approximation algorithm with running time $\mathcal{O}(k|S|)$ if we assume that $d$ can be evaluated in constant time (this is the case, e.g., if a problem instance is given via the distance matrix~$(d(s,s'))_{s,s'\in S}$).
This greedy strategy 
chooses  an arbitrary element of the data set as first center and then iteratively selects  the data point with maximum distance to the current set of~centers as the next center to be added.

We consider a fair variant of the $k$-center problem as described in Section~\ref{section_introduction}. Our variant also allows for the user to specify a  subset 
$C_0\subseteq S$ 
that {\it has to be included}
in the set of centers (think of the example of the image database and the case
that we always want to show five prespecified images
as part of the summary). Assuming that $S=\dot{\cup}_{i=1}^m {S_i}$, where $S_1, \ldots S_m$ are the $m$ demographic groups, the fair $k$-center problem 
can be stated as 
the minimization problem
\begin{align}\label{fair_k_center}
\minimize_{\substack{C=\{c_1,\ldots,c_k\}\subseteq S:\\ |C\cap S_i|=k_{S_i},\; i=1,\ldots,m}} \;\max_{s\in S} \;d(s,C\cup C_0),
\end{align}
where $k_{S_i}\in\N_0$ with $\sum_{i=1}^m k_{S_i}=k$  and $C_0\subseteq S$ are given. 
By means of a partition matroid, the fair $k$-center problem can be phrased as a matroid center problem,
for which \citet{chen_matroid_center} provide a 3-approximation algorithm 
using 
 matroid intersection 
 \citep[e.g., ][]{cook_comb_opt}. 
\citet{chen_matroid_center} do not discuss the running time of their algorithm, but  it  requires to sort all distances 
between elements in $S$ and hence has running time at least $\Omega(|S|^2\log|S|)$.
In our experiments in Section~\ref{section_experiments} we observe a 
running
 time~in~$\Omega(|S|^{5/2})$.

\section{A Linear-time Approximation Algorithm}\label{section_algorithm}

\begin{algorithm}[t!]
   \caption{Approximation algorithm for \eqref{unfair_k_center_general}}
   \label{alg_greedy_standard}
\begin{algorithmic}[1]
   \STATE {\bfseries Input:} 
   metric $d:S\times S\rightarrow \R_{\geq 0}$; 
   $k\in \N_0$; $C_0'\subseteq S$

\vspace{1mm}  
   \STATE {\bfseries Output:} $C=\{c_1,\ldots,c_k\}\subseteq S$
   
\vspace{2mm}
\STATE{set $C=\emptyset$}
\FOR{$i=1$ \TO $i=k$}
\STATE{choose $c_i\in \argmax_{s\in S} d(s,C \cup C_0')$}
\STATE{set $C=C\cup \{c_i\}$}
\ENDFOR
\RETURN $C$
\end{algorithmic}
\end{algorithm}

In this section, we present our approximation algorithm for the minimization problem \eqref{fair_k_center}. It is a recursive algorithm with respect to the number of groups~$m$. 
To increase comprehensibility, 
we first present the case of two groups and then the general case 
of an arbitrary number of groups. 

At several points, we will consider the standard 
(unfair) 
$k$-center 
problem \eqref{unfair_k_center} generalized to the case of initially given centers~$C_0'\subseteq S$, that is
\begin{align}\label{unfair_k_center_general}
\minimize_{C=\{c_1,\ldots,c_k\}\subseteq S} ~ \max_{s\in S} \;d(s,C\cup C_0').
\end{align}
We can 
adapt the greedy strategy of~\citet{gonzalez1985} for \eqref{unfair_k_center} to problem \eqref{unfair_k_center_general} while maintaining its 2-approximation 
guarantee. For the sake of completeness, we provide the algorithm 
as
 Algorithm~\ref{alg_greedy_standard}
 and state the~following~lemma:

\vspace{2mm}
\begin{lemma}\label{lemma_greedy}
Algorithm \ref{alg_greedy_standard} is a 2-approximation algorithm for 
 the unfair $k$-center problem
\eqref{unfair_k_center_general}
with running time $\mathcal{O}((k+|C_0'|) |S|)$, assuming $d$ can be evaluated in constant time.
\end{lemma}


A proof of Lemma \ref{lemma_greedy}, similar in structure to a proof in \citet[][Section 4.2]{har-peled_geometric_approx_algs} for the strategy of \citet{gonzalez1985} for problem~\eqref{unfair_k_center},
 can be found in Appendix~\ref{appendix_proofs}.

\subsection{Fair $k$-Center with Two Groups}\label{sec_approxalg_mequals2}

Assume that $S=S_1\dot{\cup} S_2$. Our algorithm first runs Algorithm \ref{alg_greedy_standard} for the unfair problem \eqref{unfair_k_center_general} with $k=k_{S_1}+k_{S_2}$ and $C_0'=C_0$. 
If we are lucky and 
Algorithm~\ref{alg_greedy_standard} 
picks $k_{S_1}$ many centers from $S_1$ and $k_{S_2}$ many centers from $S_2$, our algorithm terminates.
Otherwise, 
Algorithm~\ref{alg_greedy_standard} 
picks too many centers from one group, say $S_1$, and too few from $S_2$. 
We try to decrease the number of centers in $S_1$ by replacing any such a center with an element in its 
cluster belonging to $S_2$.
 Once we have made all such available swaps,
 the remaining clusters with centers in $S_1$ are entirely contained within $S_1$. 
 We then run Algorithm~\ref{alg_greedy_standard} on 
these 
clusters  
 with $k=k_{S_1}$ and the centers from $S_2$ as well as  $C_0$ as initially given centers, and  return both the centers from the recursive call (all in $S_1$) and those from the initial call and the swapping  in~$S_2$.

 This algorithm is formally stated 
as
Algorithm~\ref{algorithm_2groups}. 
The following theorem states that 
it 
is a 5-approximation algorithm and 
that our analysis is tight---in general,  
Algorithm~\ref{algorithm_2groups}
 does  not achieve a better approximation~factor.

\begin{algorithm}[t!]
    \caption{Approximation algorithm for \eqref{fair_k_center} when $m=2$}
   \label{algorithm_2groups}
\begin{algorithmic}[1]
      \STATE {\bfseries Input:} 
metric $d:S\times S\rightarrow \R_{\geq 0}$;       
      $k_{S_1},k_{S_2}\in \N_0$ with 
      $k_{S_1}+k_{S_2}=k$; $C_0\subseteq S$; group-membership vector~$\in \{1,2\}^{|S|}$ 
      encoding membership in $S_1$ or $S_2$

\vspace{1mm}  
   \STATE {\bfseries Output:} $C^A=\{c_1^A,\ldots,c_k^A\}\subseteq S$
   
\vspace{2mm}
\STATE{run Algorithm \ref{alg_greedy_standard} on $S$ with $k=k_{S_1}+k_{S_2}$ and $C_0'=C_0$; 
let $\widetilde{C}^A=\{\tilde{c}_1^A,\ldots,\tilde{c}_k^A\}$ denote its output}
\STATE{}
 \IF[\emph{implies} $|\widetilde{C}^A\cap S_2|=k_{S_2}$]{$|\widetilde{C}^A\cap S_1|=k_{S_1}$}
\RETURN $\widetilde{C}^A$
\ENDIF
\STATE{\#\emph{ we assume $|\widetilde{C}^A\cap S_1|>k_{S_1}$; otherwise we switch the role of $S_1$ and $S_2$}}
\vspace{1pt}
\STATE{form clusters $L_1,\ldots,L_k,L_1',\ldots,L_{|C_0|}'$ by assigning every $s\in S$ to its closest center 
in $\widetilde{C}^A\cup C_0$}
\WHILE{$|\widetilde{C}^A\cap S_1|>k_{S_1}$ \textbf{and} there exists $L_i$ with center $\tilde{c}_i^A\in S_1$ and $y\in L_i\cap S_2$}
\STATE{replace center $\tilde{c}_i^A$ with $y$ by setting $\tilde{c}_i^A=y$}
\ENDWHILE
\STATE{}
 \IF[\emph{implies} $|\widetilde{C}^A\cap S_2|=k_{S_2}$]{$|\widetilde{C}^A\cap S_1|=k_{S_1}$}
\RETURN $\widetilde{C}^A$
\ENDIF
\STATE{let $S'=\cup_{i \in[k]: \tilde{c}_i^A\in S_1}L_i$~~~~~~\# \emph{we have} $S'\subseteq S_1$}
\STATE{run Algorithm \ref{alg_greedy_standard} on $S'\cup C_0'$ with $k=k_{S_1}$ and $C_0'=C_0\cup(\widetilde{C}^A\cap S_2)$; 
let $\widehat{C}^A$ denote its output}
\vspace{2pt}
\RETURN $\widehat{C}^A\cup (\widetilde{C}^A\cap S_2)$ as well as $(k_{S_2}-|\widetilde{C}^A\cap S_2|)$ many arbitrary elements from $S_2$
\end{algorithmic}
\end{algorithm}

\vspace{2mm}
\begin{theorem}\label{theorem_2groups}
Algorithm \ref{algorithm_2groups} is a 5-approximation algorithm for the fair $k$-center problem \eqref{fair_k_center} 
with 
$m=2$, but not a $(5-\varepsilon)$-approximation algorithm for any~$\varepsilon>0$. 
It can be implemented in time $\mathcal{O}((k+|C_0|) |S|)$, assuming $d$ can be evaluated in constant time.
\end{theorem}

\vspace{-4mm}
\begin{proof}
Here we only present a sketch of the proof. The full proof can be found in Appendix~\ref{appendix_proofs}. 
For showing that  Algorithm \ref{algorithm_2groups} is a 5-approximation algorithm, let
$r^*_{\fair}$
be the optimal value of  \eqref{fair_k_center}
and 
$r^*$ 
be the optimal value of  \eqref{unfair_k_center_general} 
(for $C_0'=C_0$). 
Clearly, $r^*\leq r^*_{\fair}$. 
%
Let
$C^A$ 
 be the set of centers returned by Algorithm~\ref{algorithm_2groups}. 
 It is clear that 
 $C^A$ comprises 
 $k_{S_1}$ many elements from $S_1$ and 
$k_{S_2}$ many elements from $S_2$. 
 We need to show that 
 $\min_{c\in C^A \cup C_0} d(s,c) \leq 5r^*_{\fair}$ for every $s\in S$.
%
%
Let $\widetilde{C}^A$
be the output 
 of Algorithm~\ref{alg_greedy_standard} when called in Line 3 of Algorithm~\ref{algorithm_2groups}. 
 Since Algorithm~\ref{alg_greedy_standard} 
 is a 2-approximation algorithm for 
 \eqref{unfair_k_center_general} according to 
 Lemma~\ref{lemma_greedy}, 
 we have 
 $\min_{c\in \widetilde{C}^A \cup C_0} d(s,c) \leq 2r^*\leq 2r^*_{\fair}$, $s\in S$. 
Assume that $|\widetilde{C}^A\cap S_1|>k_{S_1}$. 
It follows from the triangle inequality  
that after exchanging centers in the while-loop in Line~9 of Algorithm~\ref{algorithm_2groups} 
we have 
$\min_{c\in \widetilde{C}^A \cup C_0} d(s,c) \leq 4r^*_{\fair}$, $s\in S$. 
Assume that still $|\widetilde{C}^A\cap S_1|>k_{S_1}$.
We 
only
need to show that 
$\min_{c\in C^A \cup C_0} d(s,c) \leq 5r^*_{\fair}$ for $s\in S'$.
Let $C^*_{\fair}$ be an optimal solution to \eqref{fair_k_center}.
We split $S'$ into two subsets $S'=S'_a\dot{\cup} S'_b$, where $S'_a$ comprises all $s\in S'$ for which the 
closest center in $C^*_{\fair}\cup C_0$ is in $S_2 \cup C_0$. Using the triangle inequality we can show that  
$\min_{c\in C^A \cup C_0} d(s,c) \leq 5r^*_{\fair}$, $s\in S'_a$. We partition $S'_b$ into at 
most $k_{S_1}$ many clusters corresponding to the closest center in $C^*_{\fair}$. Each of these clusters has 
diameter not greater than $2r^*_{\fair}$. If Algorithm~\ref{alg_greedy_standard} in Line~15 of Algorithm~\ref{algorithm_2groups} 
chooses one element from each of these clusters, we immediately have 
$\min_{c\in C^A \cup C_0} d(s,c) \leq 2r^*_{\fair}$, $s\in S'_b$. Otherwise,  Algorithm~\ref{alg_greedy_standard} 
chooses 
an 
element from $S'_a$ or two elements from the same cluster of $S'_b$. In both cases, 
it follows  from 
the greedy choice property of Algorithm~\ref{alg_greedy_standard} 
that 
$\min_{c\in C^A \cup C_0} d(s,c) \leq 5r^*_{\fair}$,~$s\in S'_b$.

A family of examples shows that Algorithm \ref{algorithm_2groups}
is not a $(5-\varepsilon)$-approximation algorithm for any~$\varepsilon>0$.
\end{proof}

\subsection{Fair $k$-Center with Arbitrary Number of Groups}\label{sec_approxalg_arbitrary_m}

The main idea to handle an arbitrary number of groups~$m$ is the same as for the case $m=2$: we first run Algorithm~\ref{alg_greedy_standard}. 
We then 
exchange centers for elements in their clusters in such a way that the number of centers from a group $S_i$ 
comes closer to $k_{S_i}$, 
which is the requested number of centers from $S_i$. 
If 
via 
exchanging centers we 
can actually hit $k_{S_i}$ for every group $S_i$, 
we are done. 
Otherwise, we wish that, when no more exchanging is possible, we are left with a subset $S'\subseteq S$  that only comprises elements 
from $m-1$ or fewer groups. Denote 
the set of 
these groups by $\mathcal{G}$. 
We also wish that for those groups not in $\mathcal{G}$ 
we have picked only the requested number of centers~or~fewer and we can consider the groups not in $\mathcal{G}$ to have been ``resolved''. 
If both are true, we can recursively apply our algorithm to $S'$ and a smaller number of groups. We might 
recurse down 
to the case of only one 
group, which we can solve with Algorithm \ref{alg_greedy_standard}.

\begin{figure}[t]
\centering
\includegraphics[scale=0.98]{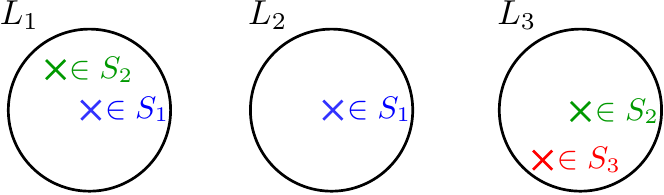}

\caption{An example illustrating the need for a more sophisticated procedure for exchanging centers in the case of three or 
more groups compared to the case of only two groups: we would like to exchange a center from $S_1$ for an element from $S_3$, 
but cannot do that directly. Rather, we have to make a series of exchanges.}\label{fig_sketch}
\end{figure}

The difficulty with this idea comes from the exchanging process. Formally, we are given $k$ centers $\tilde{c}_1^A,\ldots,\tilde{c}_k^A$ 
and the 
corresponding 
clustering $S\setminus S_{C_0}=\dot{\cup}_{i=1}^k L_i$, where $S_{C_0}=\dot{\cup}_{i=1}^{|C_0|} L_i'$ is the union of clusters with a center in $C_0$, and 
we want to exchange some centers $\tilde{c}_i^A$ for an element in their cluster $L_i$ such that there exists a strict subset of 
groups  $\mathcal{G}\subsetneq \{S_1,\ldots,S_m\}$ with the following properties:
\begin{align}
\bigcup_{i\in[k]:\, \text{$\tilde{c}_i^A$ is from a group in $\mathcal{G}$}} L_i ~\subseteq~ \bigcup_{S_i\in \mathcal{G}} S_i,\label{property_G_1}\\
\forall S_j\in\{S_1,\ldots,S_m\}\setminus \mathcal{G}: \;\sum_{i=1}^k \charfct\left\{\tilde{c}_i^A\in S_j\right\}\leq k_{S_j}.\label{property_G_2}
\end{align}
While in the case of only two groups this can easily be achieved 
by exchanging centers from the group that has more than 
the requested number of centers for elements from the other group, as we do in Algorithm~\ref{algorithm_2groups}, it is not immediately 
clear how to deal with a situation as shown in Figure \ref{fig_sketch}. 
There are three groups $S_1,S_2,S_3$ (elements of these groups are shown in blue, green, and red, respectively), 
and we have $k_{S_1}=k_{S_2}=k_{S_3}=1$. For the current set of centers (elements at the centers of the circles)  
there does not exist $\mathcal{G}\subsetneq \{S_1,S_2,S_3\}$ 
satisfying \eqref{property_G_1} and \eqref{property_G_2}. We would like to decrease the number of centers in $S_1$ and increase the 
number of centers in $S_3$, but the clusters with a center in $S_1$ do not comprise an element from $S_3$.
Hence, we cannot directly 
exchange a center from $S_1$ for an element in $S_3$. Rather, we first have to exchange a center from $S_1$ for an element in $S_2$ 
(although this increases the number of centers from $S_2$ over 
$k_{S_2}$)
and then a center from $S_2$ for an element in $S_3$. 
An algorithm that can deal with such a situation is Algorithm~\ref{alg_helper}. It exchanges some centers for an element in 
their cluster $L_i$ and yields $\mathcal{G}\subsetneq \{S_1,\ldots,S_m\}$ that provably satisfies \eqref{property_G_1} and \eqref{property_G_2}, 
as stated by the following lemma. Its proof can be found in Appendix~\ref{appendix_proofs}.

\begin{algorithm}[t!]
   \caption{Algorithm for exchanging centers \& finding~$\mathcal{G}$}
   \label{alg_helper}
\begin{algorithmic}[1]
      \STATE {\bfseries Input:} centers $\tilde{c}_1^A,\ldots,\tilde{c}_k^A$ 
and 
the 
corresponding 
clustering $S\setminus S_{C_0}=\dot{\cup}_{i=1}^k L_i$; 
$k_{S_1},\ldots,k_{S_m}\in \N_0$ with 
      $\sum_{i=1}^m k_{S_i}=k$;  
group-membership 
vector $\in\{1,\ldots,m\}^{|S\setminus S_{C_0}|}$

\vspace{1mm}  
   \STATE {\bfseries Output:} $\tilde{c}_1^A,\ldots,\tilde{c}_k^A$, where some centers $\tilde{c}_i^A$ have 
   been replaced with   
   an element  in  $L_i$, and $\mathcal{G}\subsetneq \{S_1,\ldots,S_m\}$ satisfying \eqref{property_G_1} and \eqref{property_G_2}
   
\vspace{3mm}
\STATE{set $\tilde{k}_{S_j}=\sum_{i=1}^k \charfct\left\{\tilde{c}_i^A\in S_j\right\}$ for $S_j\in\{S_1,\ldots,S_m\}$}
\vspace{-\baselineskip}
\vspace{1pt}
\STATE{construct a directed unweighted graph $G$ on $V=\{S_1,\ldots,S_m\}$ as follows: we have $S_i\rightarrow S_j$, that is 
there is a directed edge from $S_i$ to $S_j$, if and only if there exists 
 $L_t$ with center $\tilde{c}_t^A\in S_i$ and $y\in L_t\cap S_j$}
\STATE{compute all shortest paths on $G$}
\STATE{}
\WHILE{$\tilde{k}_{S_j}\neq k_{S_j}$ for some $S_j$ and there exist $S_r,S_s$ such that 
$\tilde{k}_{S_r}> k_{S_r}$ and $\tilde{k}_{S_s}< k_{S_s}$ 
and there exists a shortest path $P=S_{v_0}S_{v_1}\cdots S_{v_w}$ with $S_{v_0}=S_r$, $S_{v_w}=S_s$ that connects $S_r$ to $S_s$ in $G$}
\FOR{$l=0,\ldots,w-1$}
\STATE{find $L_t$ with center $\tilde{c}_t^A\in S_{v_l}$ and $y\in L_t\cap S_{v_{l+1}}$; 
 replace $\tilde{c}_t^A$ with $y$ by setting $\tilde{c}_t^A=y$}
\ENDFOR
\STATE{update $\tilde{k}_{S_r}=\tilde{k}_{S_r}-1$ and $\tilde{k}_{S_s}=\tilde{k}_{S_s}+1$}
\STATE{recompute $G$ and all shortest paths on $G$}
\ENDWHILE
\STATE{}
\IF{$\tilde{k}_{S_j}= k_{S_j}$ for all $S_j\in \{S_1,\ldots,S_m\}$}
\RETURN $\tilde{c}_1^A,\ldots,\tilde{c}_k^A$ and $\mathcal{G}=\emptyset$ 
\ELSE
\STATE{set 
$\mathcal{G}'=\{S_j\in\{S_1,\ldots,S_m\}: \tilde{k}_{S_j}> k_{S_j}\}$ 
and 
$\mathcal{G}=\mathcal{G}'\cup\{S_j\in\{S_1,\ldots,S_m\}\setminus \mathcal{G}':$
there exists $S_{i}\in  \mathcal{G}'$ and a path from $S_{i}$ to $S_j$ in $G\}$}
\RETURN $\tilde{c}_1^A,\ldots,\tilde{c}_k^A$ and $\mathcal{G}$
\ENDIF
\end{algorithmic}
\end{algorithm}

\vspace{2mm}
\begin{lemma}\label{lemma_alg_helper}
Algorithm \ref{alg_helper} is well-defined, it terminates, and exchanges centers in such a way that the set~$\mathcal{G}$ that it returns satisfies 
$\mathcal{G}\subsetneq \{S_1,\ldots,S_m\}$
and properties~\eqref{property_G_1} and \eqref{property_G_2}.
\end{lemma}

Observing that the number of iterations of the 
 while-loop 
 in Line~7 
 is upper-bounded by $k$ as 
 the proof of Lemma \ref{lemma_alg_helper} shows, 
 that the number of iterations of the 
 for-loop in Line~8 is upper-bounded by $m$, and that all shortest paths on $G$ can be computed in running time $\mathcal{O}(m^3)$ 
 \citep[][Chapter~25]{cormen_book}, 
 it is not hard to see  
 that Algorithm \ref{alg_helper} can be implemented with running time 
 $\mathcal{O}(k m |S|+km^3)$.

 Using Algorithm \ref{alg_helper}, it is straightforward to design a recursive 
 approximation algorithm for the fair
 $k$-center problem~\eqref{fair_k_center} as outlined at the beginning of Section~\ref{sec_approxalg_arbitrary_m}. 
 We 
 state 
 the 
 algorithm as Algorithm~\ref{algorithm_mgroups}. Applying, 
 by means of induction,  
 a similar technique 
 as in the proof of Theorem~\ref{theorem_2groups} 
  to every (recursive) call of 
 Algorithm~\ref{algorithm_mgroups}, we can prove the following:

 \begin{algorithm}[t!]
   \caption{Approximation alg. for \eqref{fair_k_center} for arbitrary $m$}
   \label{algorithm_mgroups}
\begin{algorithmic}[1]
    \STATE   {\bfseries Input:} 
      metric $d:S\times S\rightarrow \R_{\geq 0}$; 
      $k_{S_1},\ldots,k_{S_m}\in \N_0$ with 
      $\sum_{i=1}^m k_{S_i}=k$; $C_0\subseteq S$; group-membership vector $\in \{1,\ldots,m\}^{|S|}$

\vspace{1mm}  
   \STATE {\bfseries Output:} $C^A=\{c_1^A,\ldots,c_k^A\}\subseteq S$
   
\vspace{2mm}
\STATE{run Algorithm \ref{alg_greedy_standard} on $S$ with $k=\sum_{i=1}^m k_{S_i}$ and \\$C_0'=C_0$; 
let $\widetilde{C}^A=\{\tilde{c}_1^A,\ldots,\tilde{c}_k^A\}$ denote its output}
\vspace{1pt}
 \IF{$m=1$}
\RETURN $\widetilde{C}^A$
\ENDIF
\STATE{}
\STATE{form clusters $L_1,\ldots,L_k,L_1',\ldots,L_{|C_0|}'$ by assigning every $s\in S$ to its closest center 
in $\widetilde{C}^A\cup C_0$}
\STATE{apply Algorithm \ref{alg_helper} to $\tilde{c}_1^A,\ldots,\tilde{c}_k^A$ 
and  $\dot{\cup}_{i=1}^k L_i$ in order to exchange some 
centers 
$\tilde{c}_i^A$ and obtain $\mathcal{G}\subsetneq \{S_1,\ldots,S_m\}$}
 \IF{$\mathcal{G}=\emptyset$}
\RETURN $\widetilde{C}^A$
\ENDIF
\STATE{}
\STATE{let 
$S'=\cup_{i\in[k]:\, \text{$\tilde{c}_i^A$ is from a group in $\mathcal{G}$}}\, L_i$ and\\ 
$C'=\{\tilde{c}_i^A\in \widetilde{C}^A:\tilde{c}_i^A\text{ is from a group not in }\mathcal{G}\}$; 
recursively call Algorithm \ref{algorithm_mgroups}, where: 
\vspace{-4pt}
\begin{itemize}[leftmargin=*]
\setlength{\itemsep}{-2pt} \item 
$S'\cup C'\cup C_0$ plays the role of $S$
\item we assign elements in $C'\cup C_0$ 
to an arbitrary group in $\mathcal{G}$ and hence there are $|\mathcal{G}|<m$ many groups $S_{j_1},\ldots,S_{j_{|\mathcal{G}|}}$
\item the requested numbers of centers are $k_{S_{j_1}},\ldots,k_{S_{j_{|\mathcal{G}|}}}$
\item $C'\cup C_0$ plays the role of initially given centers~$C_0$
\end{itemize}
let $\widehat{C}^R$ denote its output}
\vspace{2pt}
\RETURN $\widehat{C}^R\cup C'$ as well as $(k_{S_j}-|C'\cap S_j|)$ many arbitrary elements from $S_j$ for every group $S_j$ not in $\mathcal{G}$
\end{algorithmic}
\end{algorithm}

 \vspace{2mm}
\begin{theorem}\label{theorem_mgroups}
Algorithm \ref{algorithm_mgroups} is a $(3\cdot 2^{m-1}-1)$-approximation algorithm for the fair 
$k$-center problem~\eqref{fair_k_center} 
with 
$m$ groups. It can be implemented in time  $\mathcal{O}((|C_0|m+k m^2) |S|+km^4)$, assuming $d$ can be evaluated in constant time.
\end{theorem}
 

It is not clear to us  whether our analysis of Algorithm~\ref{algorithm_mgroups} is tight 
and the approximation factor achieved by Algorithm~\ref{algorithm_mgroups} can  indeed be as large as $(3\cdot 2^{m-1}-1)$ or whether the dependence on $m$ is actually less severe  
(compare with 
Section~\ref{section_experiments} and Section~\ref{section_discussion}).  Although trying hard to find 
instances for which the approximation factor of  Algorithm~\ref{algorithm_mgroups} is large, we never observed 
a 
factor greater than~$8$.

 \vspace{2mm}
\begin{lemma}\label{mgroups_lower_bound}
Algorithm \ref{algorithm_mgroups} is not a $(8-\varepsilon)$-approximation algorithm for 
any $\varepsilon>0$
for 
\eqref{fair_k_center} 
with 
$m\geq 3$ groups.
\end{lemma}
 
The 
proofs of Theorem  \ref{theorem_mgroups} 
and Lemma \ref{mgroups_lower_bound}
are 
in Appendix~\ref{appendix_proofs}.  

\section{Related Work}\label{section_related_work}

\textbf{Fairness~~}
By now, there is a huge body of work on fairness in machine learning. For a recent 
paper providing 
an 
overview 
of the literature on fair classification 
see \citet{donini2018}.  
 Our paper adds to the literature on fair methods 
 for unsupervised learning tasks 
\citep
{fair_clustering_Nips2017,celis_multiwinner,celis2018,celis_fair_ranking,samira2018,sohler_kmeans}. 
Note that all these papers assume
to know 
which demographic group a data point belongs to just as we do. 
We discuss the two 
 works 
 most
closely
 related to our~paper.

First, ~\citet{celis2018} also deal with   
the problem of~fair data summarization. 
They 
study the same fairness constraint 
as we do, that is the summary 
must 
contain $k_{S_i}$ many 
elements 
from group $S_i$. However, while we aim for a \emph{representative}  
summary, where every data point should be close to at least one center in the summary,  \citeauthor{celis2018} aim for a \emph{diverse}  
summary. Their approach requires the data set $S$ to 
consist of points in 
$\R^n$, and then the diversity of a subset of $S$ is measured by the volume of the parallelepiped that it spans \citep{taskar2012}. 
This summarization objective 
is different from ours, and in different 
applications
one or the other may be more appropriate. 
An advantage of our approach is that it only requires access to a metric on the data set rather than 
feature representations of 
data points.

The second line of work we 
discuss
centers
around 
the paper of~\citet{fair_clustering_Nips2017}.
Their 
 paper
proposes a notion of fairness for clustering different from ours. Based on the fairness notion of disparate impact 
\citep{feldman2015} / the $p\%$-rule \citep{zafar2017} for classification,  
the paper by \citeauthor{fair_clustering_Nips2017} asks that every group be
approximately 
equally represented in each cluster. 
In their paper, 
\citeauthor{fair_clustering_Nips2017} focus on $k$-medoid and
$k$-center clustering and the case of two groups. 
Subsequently, \citet{roesner2018} study such a fair $k$-center problem for multiple  groups, and
\citet{sohler_kmeans} build upon the work of \citeauthor{fair_clustering_Nips2017} to devise  algorithms for such a  fair $k$-means problem. 
\citet{fair_SC_2019} incorporate the fairness notion of~\citeauthor{fair_clustering_Nips2017} into the spectral clustering framework. 
While we 
certainly  
consider the fairness notion of~\citet{fair_clustering_Nips2017}, which can be applied to any kind of clustering, to be meaningful in some scenarios, we believe that in certain applications of  centroid-based clustering (such as data summarization) our proposed fairness notion provides a more 
sensible 
 alternative.
 
 \vspace{1pt}
\textbf{Centroid-based clustering~~}
There 
are many 
papers proposing heuristics and approximation algorithms for 
both $k$-center \citep[e.g., ][]{hochbaum1986,mladenovic2003,ferone2017} and $k$-medoid \citep[e.g.,][]{charikar2002,arya2004,li2013} under 
various assumptions on $S$ and the distance function~$d$. There are also numerous papers on versions 
with
 constraints, such as lower or upper bounds on the size of the  clusters  
\citep{aggarwal2010,cygan2012, roesner2018}. 
 
Most important to mention are the works by  \citet{red_blue_median}, \citet{matroid_median} and \citet{chen_matroid_center}.
 \citeauthor{red_blue_median} are the first that consider our fairness constraint 
 (for two groups)
  for $k$-medoid. They present a local search algorithm and prove it to be a constant-factor 
 approximation  algorithm. Their work  has been generalized by \citeauthor{matroid_median}, who consider $k$-medoid under the constraint 
 that the 
centers have  
 to  form an independent set in a given matroid. This kind of constraint contains~our~fairness constraint as a special case (for an arbitrary number of groups).
  \citeauthor{matroid_median} obtain a 16-approximation~algorithm for this so-called matroid median problem  
 based on rounding the solution of a linear programming relaxation. Subsequently,  \citeauthor{chen_matroid_center} study the matroid center problem. Using 
 matroid intersection as black box, they obtain a 3-approximation algorithm. 
 Note that none of  
  \citeauthor{red_blue_median}, \citeauthor{matroid_median} or \citeauthor{chen_matroid_center} discuss the running time 
  of their algorithm, except for arguing it to be polynomial 
  (see Section~\ref{section_prob_statement}).
We also mention the works by \citet{Chakrabarty2018}, who provide a generalization of the matroid center problem and in doing so recover the result of  \citet{chen_matroid_center}, and by \citet{Kale2018}, who studies the matroid center problem in a streaming~setting.

\section{Experiments}\label{section_experiments}

In this section, we present a number of experiments\footnote{Python code  
is available on
 \url{https://github.com/matthklein/fair_k_center_clustering}.}. We begin with a motivating example on a small image data set illustrating that a summary produced by Algorithm~\ref{alg_greedy_standard} (i.e., the standard greedy strategy for the unfair $k$-center problem) can 
 be quite unfair. 
We also compare summaries produced by our algorithm to summaries produced by the method of \citet{celis2018}. 
We then investigate the approximation factor 
of our algorithm
on several artificial~instances 
with known or computable  
optimal value of the fair $k$-center problem~\eqref{fair_k_center} 
and compare our algorithm to the 
one
for the matroid center problem by \citet{chen_matroid_center},  both in terms of 
approximation  factor / cost of output and running time.  
Next, on both synthetic and real data, we compare our algorithm in terms of the cost of 
its output 
 to two baseline heuristics 
 (with running time linear in $|S|$ and $k$ just as for our algorithm). 
Finally, we compare our algorithm to  
Algorithm~\ref{alg_greedy_standard} more systematically. 
We study the difference in the costs of the 
outputs of 
our algorithm and 
Algorithm~\ref{alg_greedy_standard}, 
a quantity one may refer to as \emph{price of fairness}, and measure how unfair 
the output of 
Algorithm~\ref{alg_greedy_standard} 
can be. In the following, all boxplots show 
results of 200 runs of an~experiment.

\subsection{Motivating Example and Comparison with \citet{celis2018}}

\begin{figure}[t]
\centering
\includegraphics[width=\columnwidth]{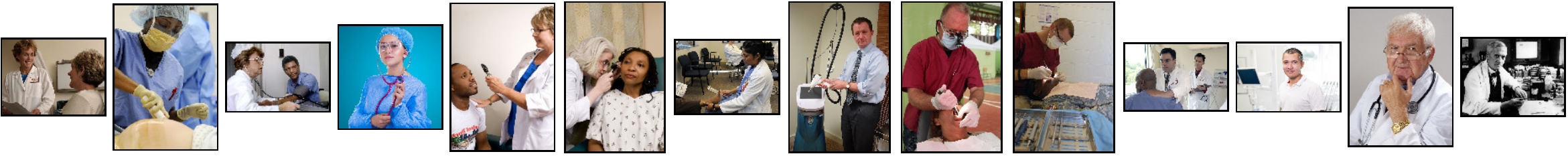}

\vspace{4mm}
\begin{tabular}{c|c|c}
Algorithm \ref{alg_greedy_standard} & Our Algorithm & \citeauthor{celis2018}\\
\hline
\includegraphics[width=0.28\columnwidth,trim={0 0 0 -0.5cm}]{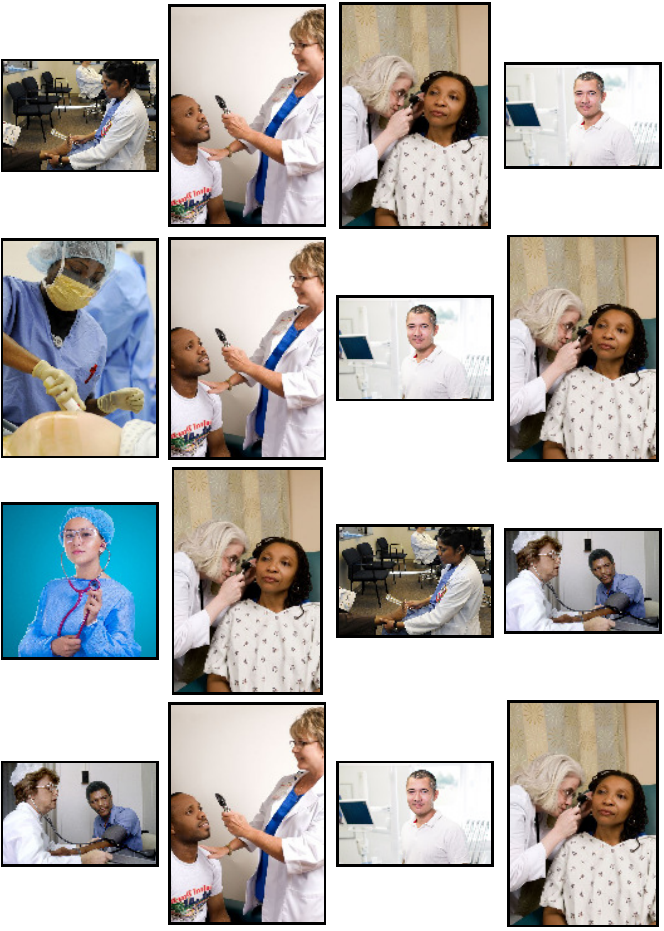}
&
\includegraphics[width=0.28\columnwidth]{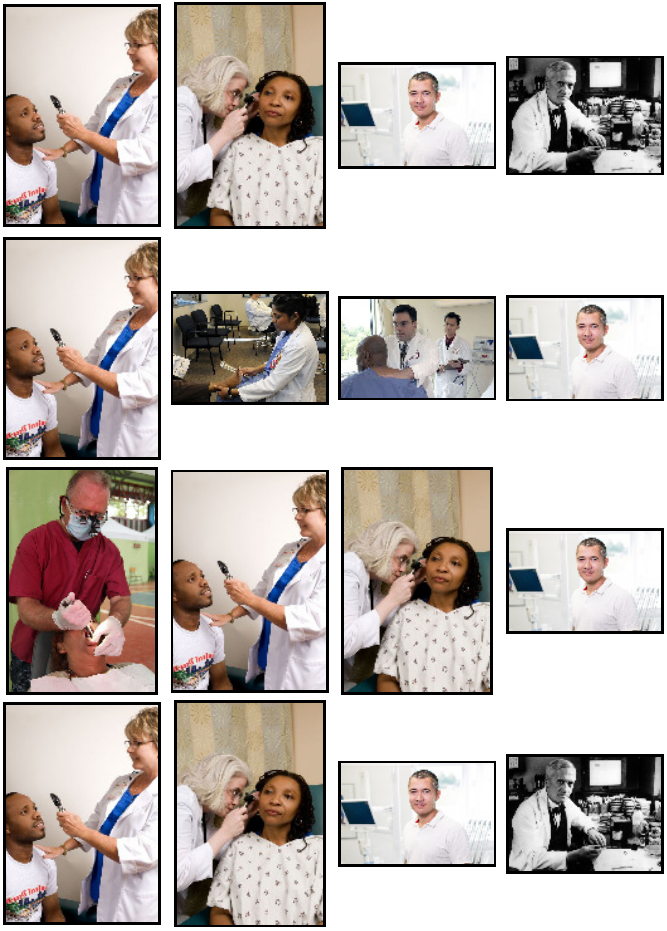}
&
\includegraphics[width=0.28\columnwidth]{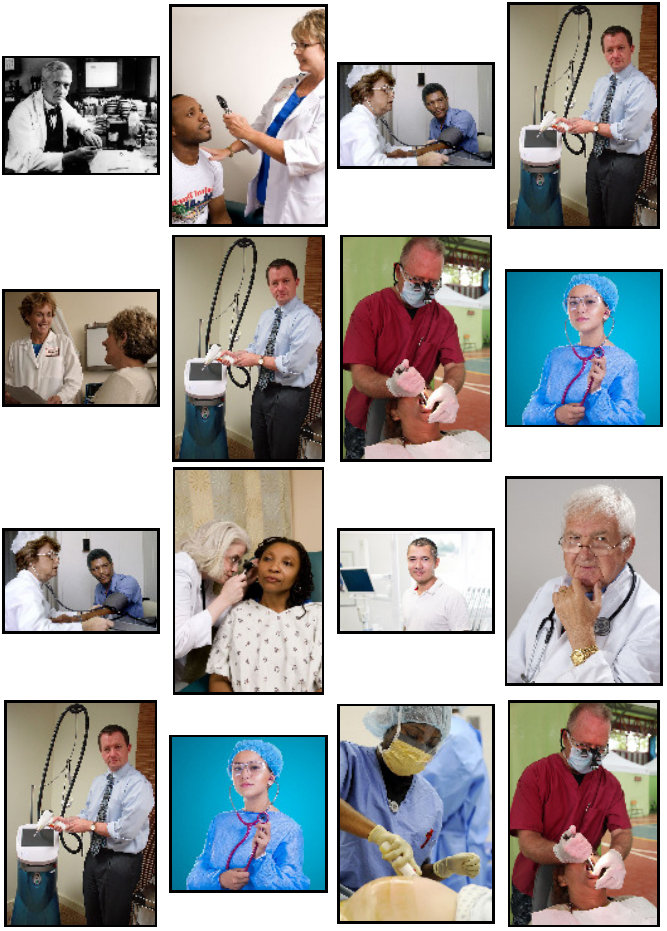}
\end{tabular}
\caption{A data set consisting of 14 images of medical doctors (7 female, 7 male) and four summaries computed by the unfair Algorithm \ref{alg_greedy_standard}, our algorithm and the algorithm proposed by \citet{celis2018} (all three algorithms are randomized algorithms).}\label{fig_exp_images}
\end{figure}

Consider the 14 images\footnote{All images were found on \url{https://pexels.com},  \url{https://pixnio.com} or \url{https://commons.wikimedia.org}  and 
are in
the public domain.} 
of medical doctors shown in the first row of Figure~\ref{fig_exp_images}. Assume we want to generate a summary of size four of these images. One way to do so is to run Algorithm~\ref{alg_greedy_standard}. 
The first column of the table in Figure~\ref{fig_exp_images} shows in each row the summary produced in one run of  Algorithm~\ref{alg_greedy_standard} (recall that all algorithms considered here are randomized algorithms). These summaries are quite unfair: although there is an equal number of images of female doctors and images of male doctors, all these summaries show three or even four females. To overcome this bias we can apply our 
algorithm or the method of \citet{celis2018}, which both allow us to explicitly state 
the numbers of females and males that we want in the summary. 
The second and the third column of the table show summaries produced by these algorithms. It is hard to say which of them produces more useful summaries and the results ultimately depend on the feature representations of the images (see the next paragraph).  To provide further illustration, we present a similar experiment in Figure~\ref{fig_exp_images_SUPPMAT} in Appendix~\ref{appendix_experiments}.

\begin{figure*}[t]
\centering
\includegraphics[scale=1.04]{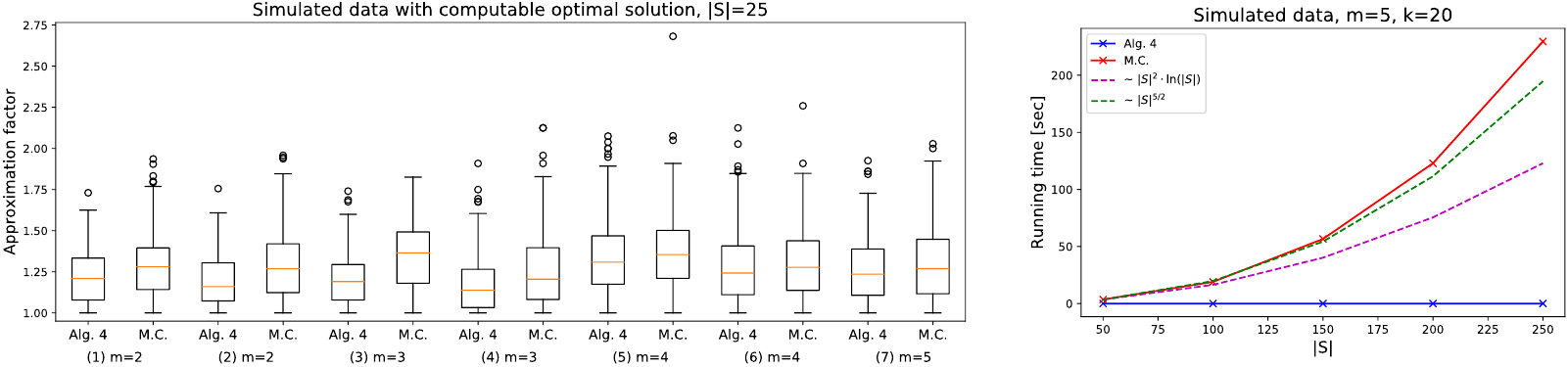}
\caption{
\textbf{Left:} 
Approx.
factor of Alg.~\ref{algorithm_mgroups} and the algorithm by \citeauthor{chen_matroid_center} (M.C.) on simulated data with computable 
optimal solution. $|S|=25$; various settings with  $m\in \{2,3,4,5\}$. 
\textbf{(1)}  $|C_0|=2$, $(k_{S_1},k_{S_2})=(2,2)$ 
%
\textbf{(2)} $|C_0|=2$, $(k_{S_1},k_{S_2})=(4,2)$ 
%
\textbf{(3)} $|C_0|=2$, $(k_{S_1},k_{S_2},k_{S_3})=(2,2,2)$
\textbf{(4)}  $|C_0|=1$, $(k_{S_1},k_{S_2},k_{S_3})=(5,1,1)$
%
\textbf{(5)}  
$C_0=\emptyset$, $(k_{S_1},k_{S_2},k_{S_3},k_{S_4})=(2,2,2,2)$
\textbf{(6)}  
$C_0=\emptyset$, 
$(k_{S_1},k_{S_2},k_{S_3},k_{S_4})=(3,3,1,1)$
\textbf{(7)} 
$C_0=\emptyset$, 
 $(k_{S_1},k_{S_2},k_{S_3},k_{S_4},k_{S_5})=(2,2,2,1,1)$.  \textbf{Right:} Running time as a function~of~$|S|$. 
}\label{fig_exp_comparison_matroid_k_center}
\end{figure*}

\begin{figure*}[t]
\hspace{0mm}
\begin{minipage}{4.5cm}
\includegraphics[scale=0.35]{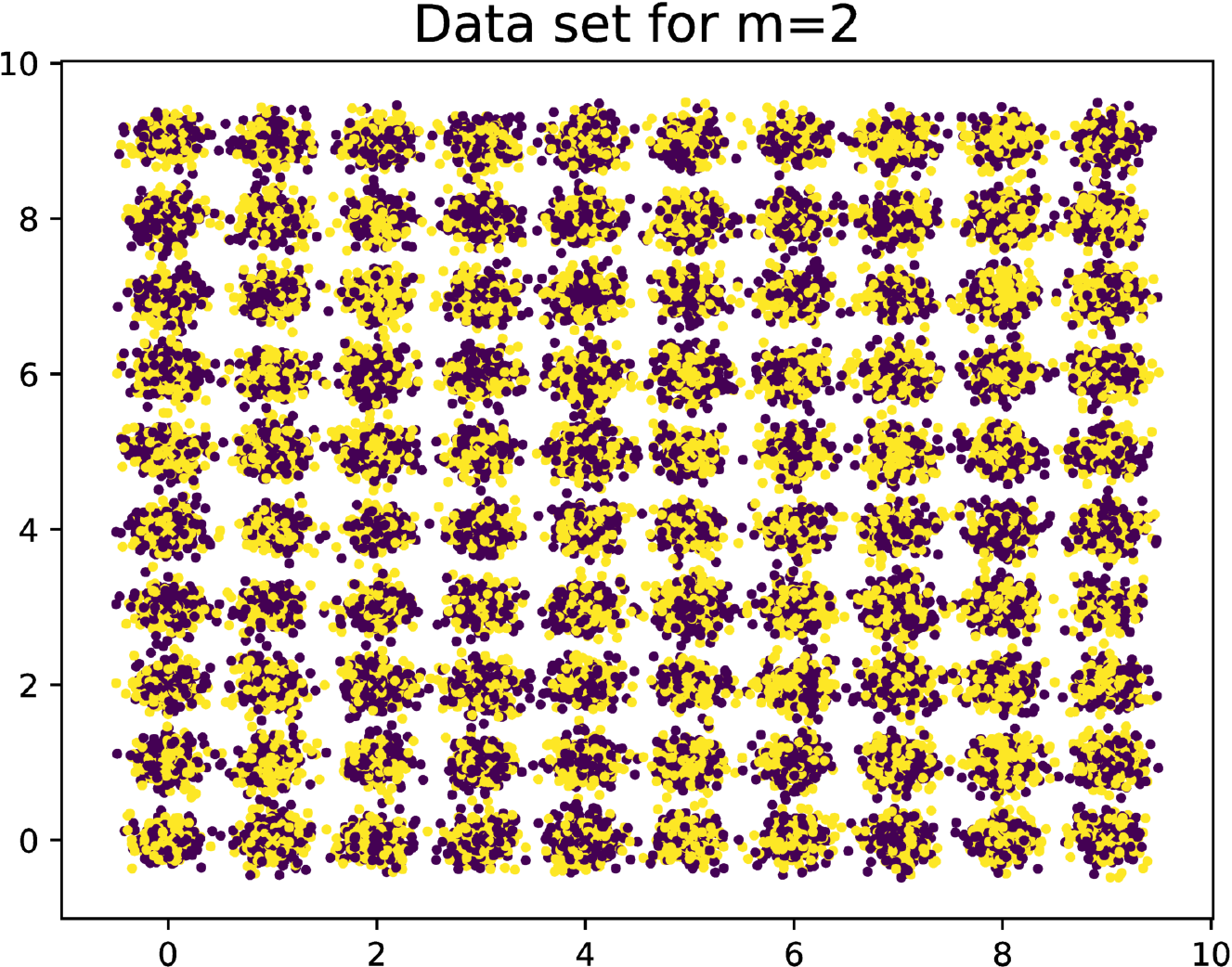}
\end{minipage}
\hspace{8mm}
\begin{minipage}{8cm}
\includegraphics[scale=0.35]{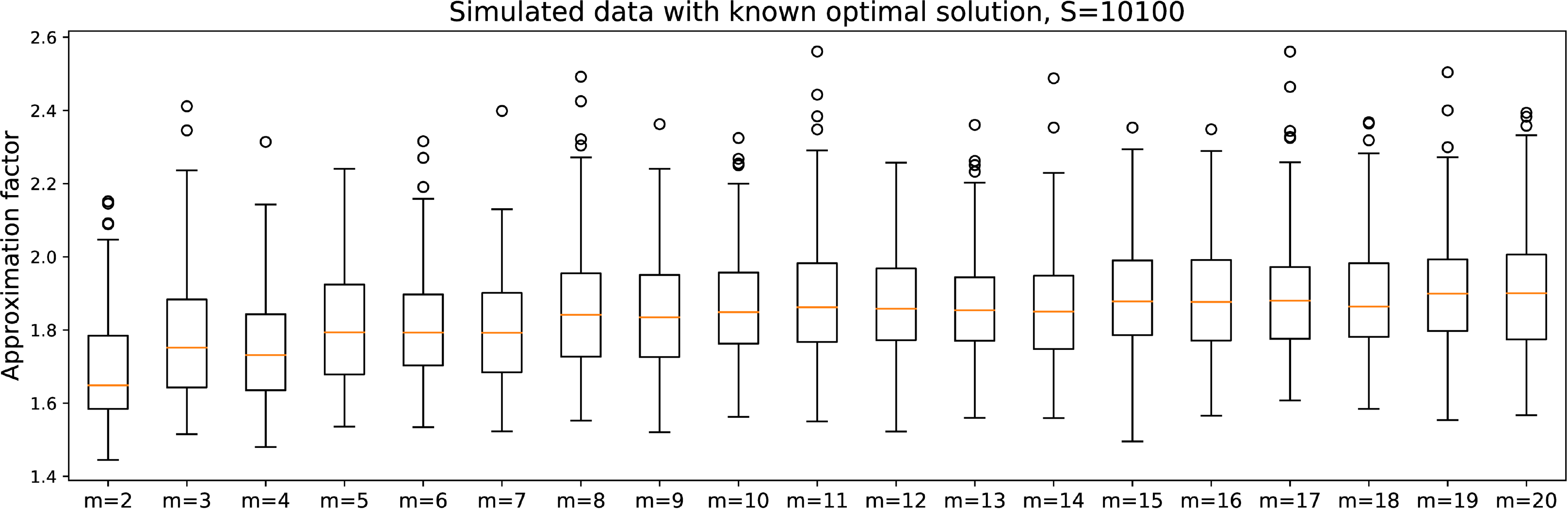}
\end{minipage}
\caption{
Approximation factor of our algorithm on simulated data with known optimal solution. $|S|=10100$, $C_0=\emptyset$, $\sum_{i=1}^m k_{S_i}=100$. \textbf{Left:} Example of the data set when $m=2$. The optimal solution consists of 100 points located at the centers of the visible clusters and has cost $0.5$.  \textbf{Right:} Approximation factor for  $m\in \{2,\ldots,20\}$. 
}\label{fig_exp_artificial_points}
\end{figure*}

For computing feature representations of the images and running the algorithm of \citeauthor{celis2018} we used the code 
provided by them.
The feature vector of an image is a 
histogram based on the image's SIFT descriptors;
 see \citeauthor{celis2018} for details. We used the Euclidean metric between these feature vectors as metric~$d$ for Algorithm~\ref{alg_greedy_standard} and our~algorithm.

\subsection{
Approximation Factor and Comparison with \citet{chen_matroid_center}
}


We implemented the algorithm by \citet{chen_matroid_center}  using the generic algorithm for matroid intersection provided in SageMath\footnote{\url{http://sagemath.org/}}. To speed up computation, rather than testing all distance values  as threshold as suggested by \citeauthor{chen_matroid_center}, we implemented binary search to look for the optimal value.

In the experiment 
shown in the left part of 
Figure \ref{fig_exp_comparison_matroid_k_center}, we study the approximation factor achieved by our algorithm (Alg.~\ref{algorithm_mgroups}) and the algorithm by \citeauthor{chen_matroid_center} (M.C.) in various settings of 
values of 
$m$, $|C_0|$ and $k_{S_i}$, $i\in[m]$. 
The data set~$S$ always consists of 25 vertices of a random graph and is small enough to explicitly compute an optimal solution to the fair $k$-center problem~\eqref{fair_k_center}. The random graph is 
constructed according to an Erd\H os-R\'enyi model, where any possible edge between two vertices is  contained in the graph with probability $2\log(|S|)/|S|$.
With high probability such a graph is connected 
(if not, 
we discard it). We put random weights on the edges, drawn from the uniform distribution on $[100]$, and let the metric~$d$ be the shortest-path distance on the graph. We assign every vertex to one of $m$ groups uniformly at random and 
randomly 
choose a subset $C_0\subseteq S$ of initially given centers.  
As we
can see from the boxplots, the approximation factor achieved by our algorithm is \emph{never} larger than~2.2. 
We also see
 that 
in each of the seven settings that we consider the median of the achieved approximation factors (indicated by the red lines 
in the 
boxes) is smaller for our algorithm than for the algorithm by \citeauthor{chen_matroid_center}.

In the experiment 
shown in the right part of 
Figure \ref{fig_exp_comparison_matroid_k_center}, we study the running time of the two algorithms as a function of the size of the data set, which is created analogously to the experiment in the left part. We set $m=5$, $C_0=\emptyset$ and $k_{S_i}=4$, $i\in[5]$. The shown curves are obtained from averaging the running times of 200 runs of the experiment (performed on an  iMac with  3.4  GHz i5 / 8 GB DDR4).  
  While our algorithm never runs for more than 0.01 seconds, the algorithm by \citeauthor{chen_matroid_center}, on average, runs for 230 seconds when $|S|=250$. 
Its 
run 
time grows at least as $|S|^{5/2}$, which proves it to be inappropriate for massive data sets.
Boxplots of the costs of the outputs 
obtained
 in this experiment are provided in Figure~\ref{fig_exp_cost_comp_matr_intersection} in Appendix~\ref{appendix_experiments}. We can see there that  the costs are very similar for the two algorithms.

In the experiment 
of 
Figure \ref{fig_exp_artificial_points}, 
we once more study the approximation factor achieved by our algorithm. 
We place 100 optimal centers at $(i,j)\in \R^2$, $i,j\in\{0,\ldots,9\}$, and sample $10000$ points around them such that for every center the farthest point in its cluster is at distance~0.5 from the center (Euclidean distance). One such a point set can be seen in the left plot of Figure~\ref{fig_exp_artificial_points}. We randomly assign every point and center to one of $m$ groups and set $k_{S_i}$ to the number of centers that have been assigned to group $S_i$. We let $C_0=\emptyset$. For $m\in \{2,\ldots,20\}$, the right part of Figure~\ref{fig_exp_artificial_points} shows boxplots of the approximation factors 
for our algorithm. 
Similarly as before, 
the approximation factor achieved by our algorithm is \emph{never} larger than~2.6.  Most interestingly, the approximation factor increases 
very moderately with~$m$.

\begin{figure*}[t]
\centering
\includegraphics[scale=0.3]{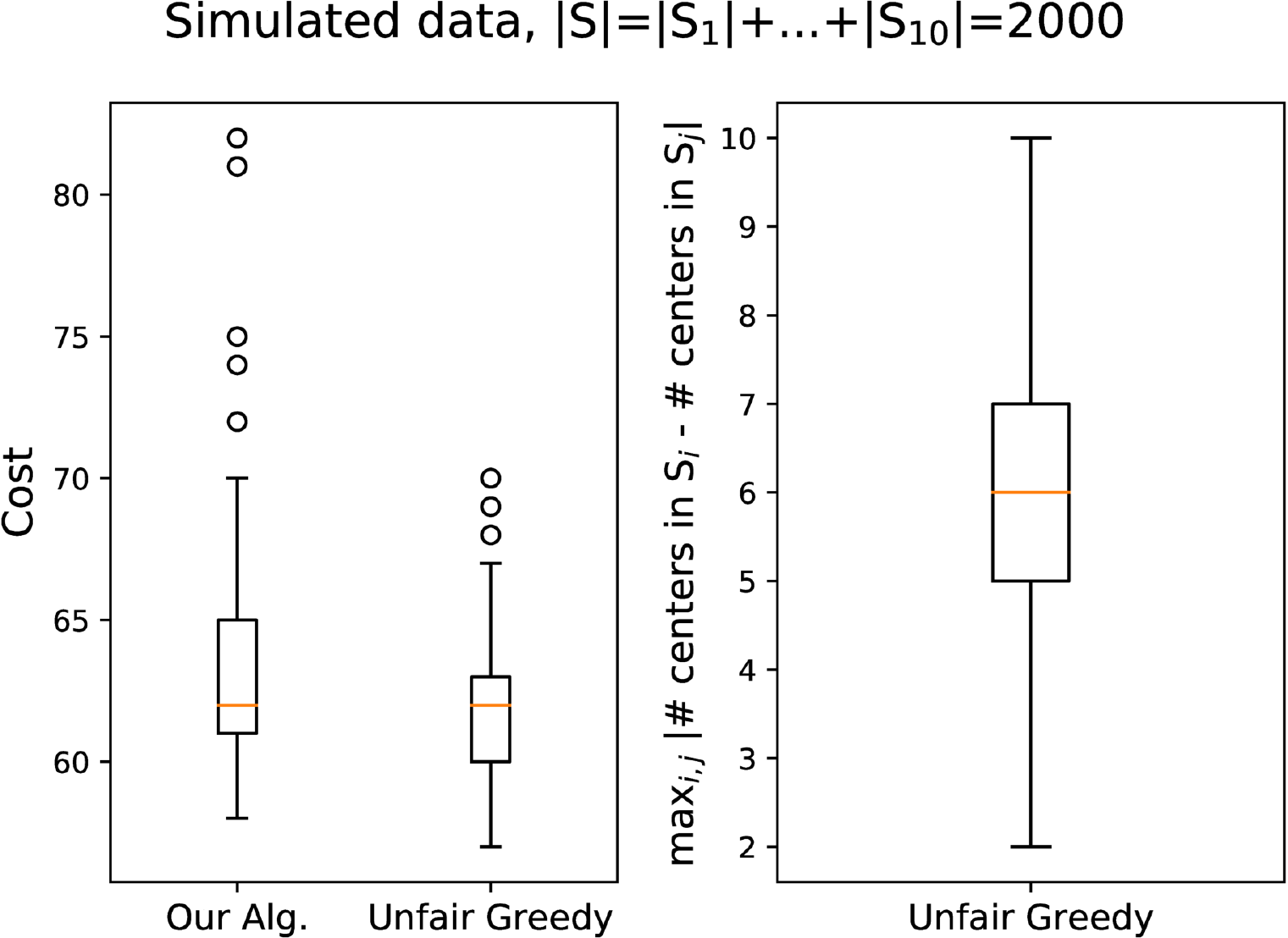}
\hspace{7mm}
\includegraphics[scale=0.3]{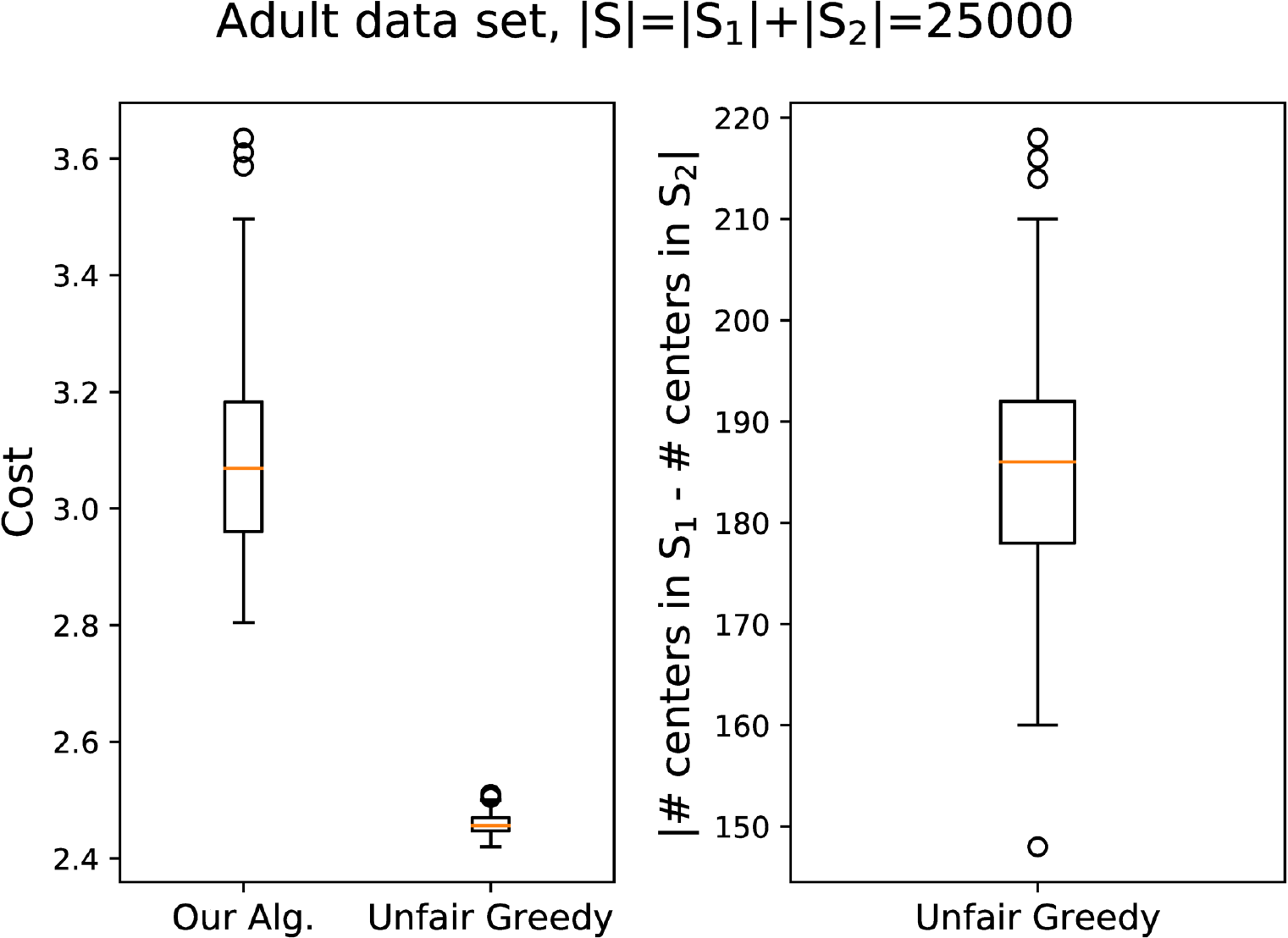}
\hspace{7mm}
\includegraphics[scale=0.3]{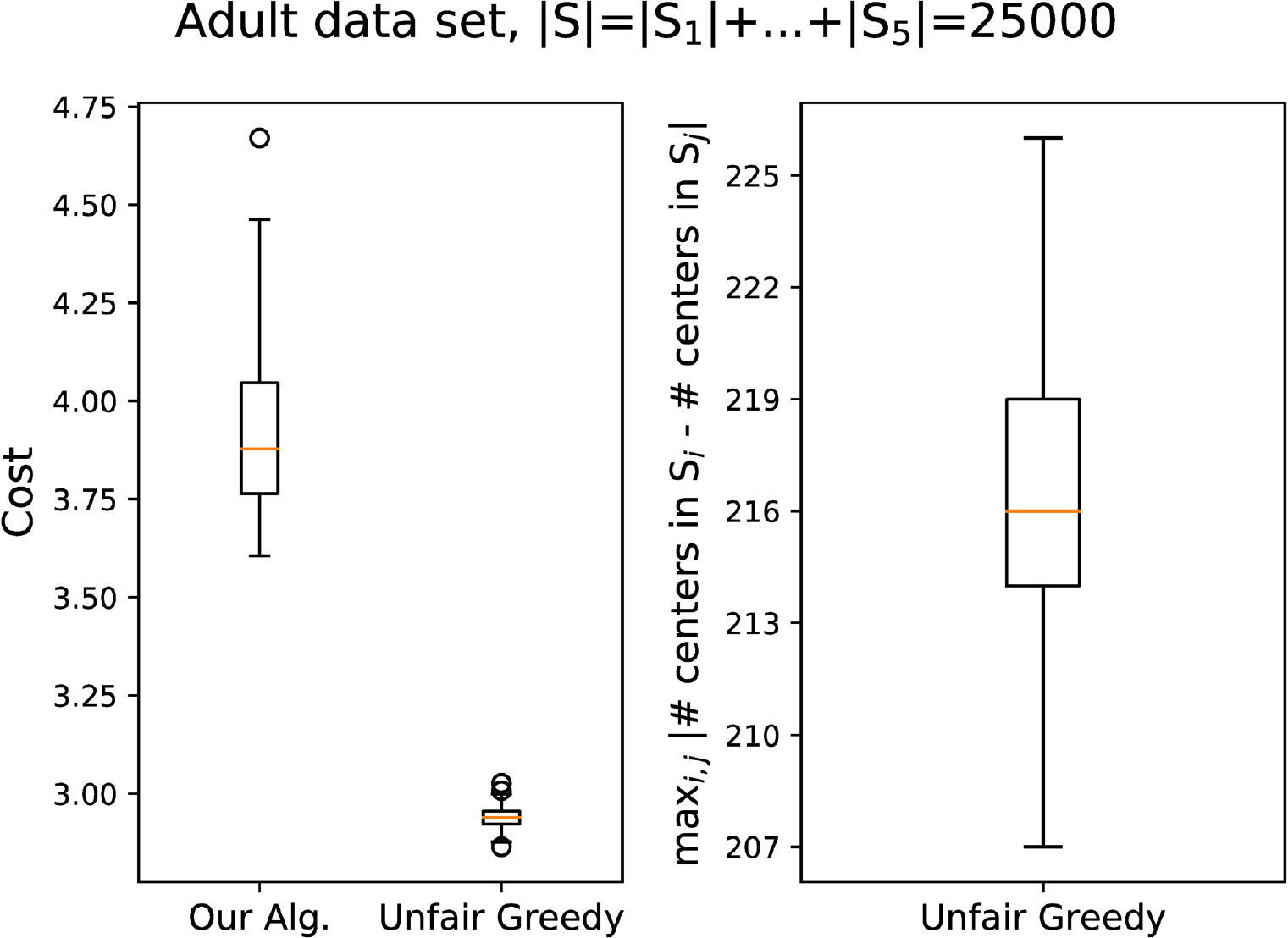}
\caption{
Cost of the 
output of
our algorithm in comparison to the unfair Algorithm~\ref{alg_greedy_standard} 
and maximum deviation of the 
numbers of centers in $S_i$ and $S_j$, $i,j\in[m]$, 
in the output of Algorithm \ref{alg_greedy_standard} 
(it is $k_{S_i}=k_{S_j}$, $i,j\in[m]$). 
\textbf{Left:} $m=10$. \textbf{Middle:}  $m=2$. \textbf{Right:} $m=5$.
}\label{fig_comp_greedy}
\end{figure*}

\subsection{
Comparison with Baseline Approaches
}

We compare our algorithm in terms of the cost of an approximate solution to two 
linear-time 
baseline 
heuristics 
for 
the fair $k$-center problem~\eqref{fair_k_center}. 
The first one, referred to as
Heuristic~A, runs Algorithm~\ref{alg_greedy_standard} on each group separately (with $k=k_{S_i}$ and $C_0'=S_i\cap C_0$ for group~$S_i$) and 
outputs the union of the centers obtained for the groups.
 The second one, 
Heuristic~B, greedily chooses centers similarly to Algorithm~\ref{alg_greedy_standard}, but only from those groups for which we have not reached the requested number of centers yet. It is easy to see that the approximation factor achieved by these 
heuristics can be arbitrarily 
large 
 on 
some worst-case~instances.

\begin{figure}[t]
\centering
\includegraphics[scale=0.29]{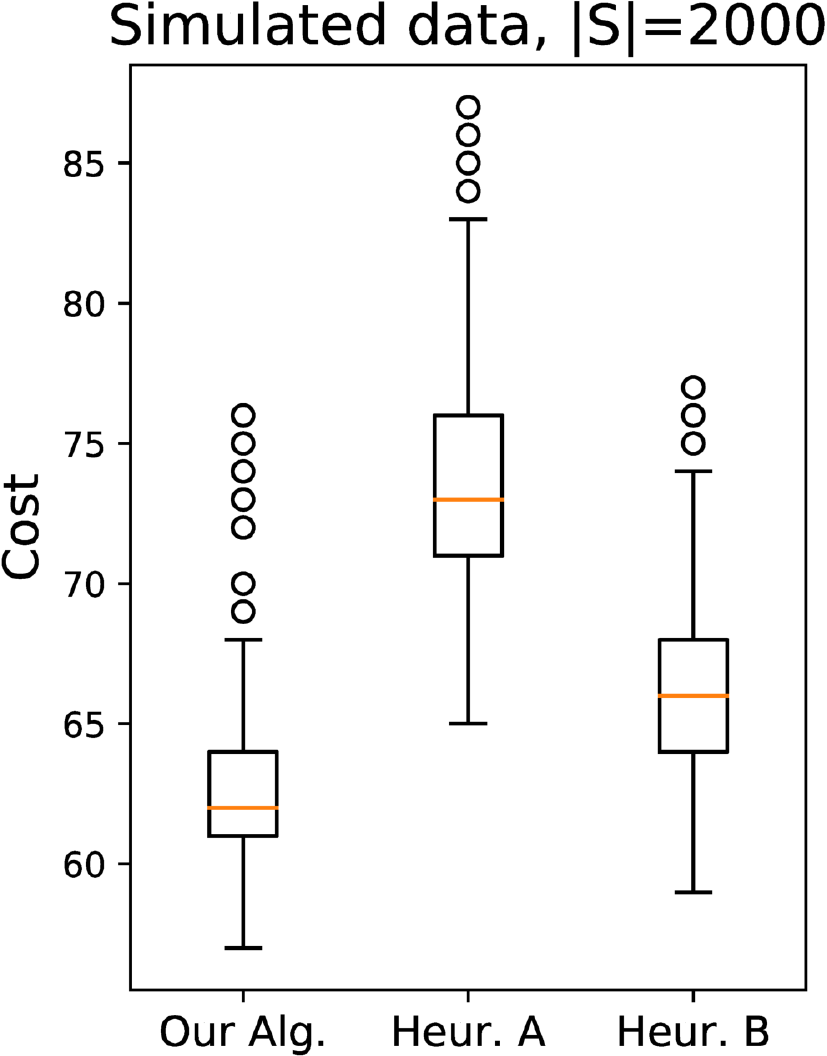}
\hspace{1mm}
\includegraphics[scale=0.29]{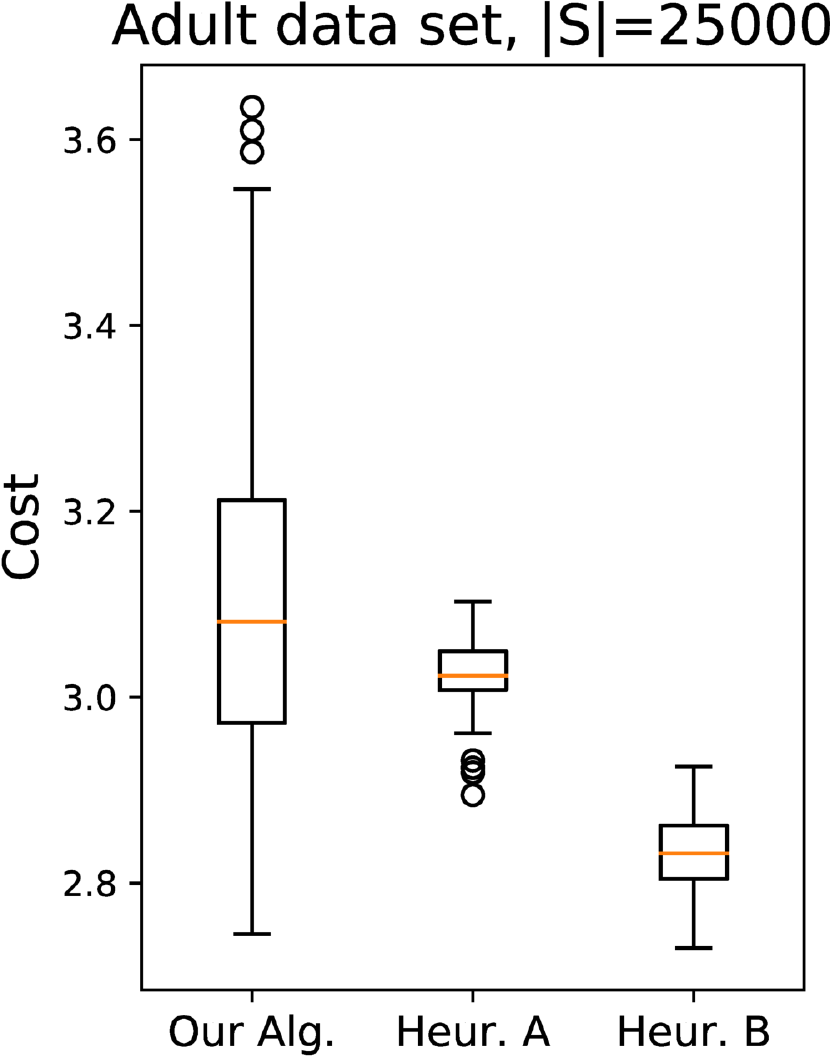}
\hspace{1mm}
\includegraphics[scale=0.29]{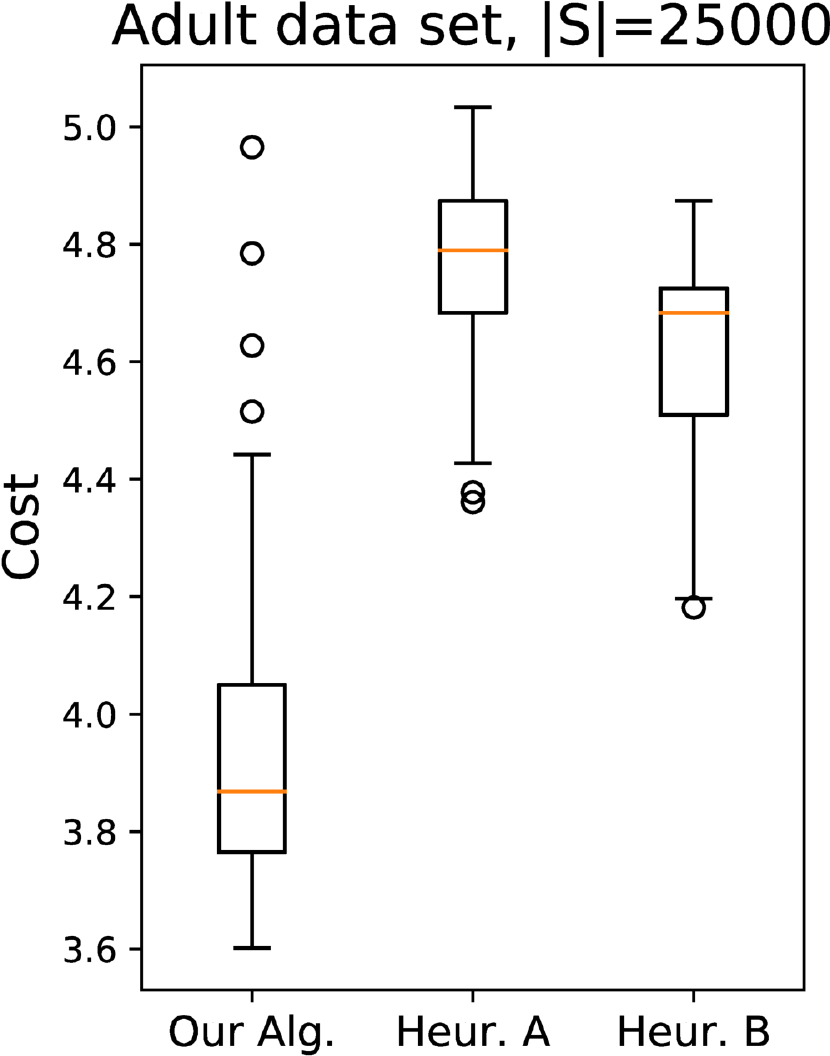}
\caption{
Cost of the 
output of 
our algorithm in comparison to 
two 
heuristics. 
\textbf{Left:} $m=10$. \textbf{Middle:}  $m=2$. \textbf{Right:} $m=5$.
}\label{fig_comp_heuristics}
\end{figure}

Figure \ref{fig_comp_heuristics} shows boxplots of the costs of the approximate solutions returned by our algorithm and the two heuristics for three data sets: the data set in the left plot consists of 
$2000$
vertices of a random graph constructed similarly as in  the experiments
of Figure~\ref{fig_exp_comparison_matroid_k_center}.
We set  $m=10$, $k_{S_i}=4$, $i\in[10]$, and $|C_0|=10$. The data set in the middle and in the right plot consists of the first 25000 records of the Adult data set \citep{Dua_2019}. We only use its
six 
numerical features (e.g., age, hours worked per week), normalized to 
zero mean and unit variance, 
for representing records and use the $l_1$-distance as metric $d$. For the experiment shown in the middle plot, we split the data set into two groups according to the sensitive feature gender 
(\#Female=8291, \#Male=16709)
and set $k_{S_1}=k_{S_2}=200$.  
For the experiment shown in the right plot, we split the data set into five groups according to the 
feature race 
(\#White=21391, \#Asian-Pac-Islander=775, \#Amer-Indian-Eskimo=241, \#Other=214, \#Black=2379) and set $k_{S_i}=50$, $i\in[5]$. 
In Figure \ref{fig_comp_heuristics_SUPPMAT} in Appendix~\ref{appendix_experiments} we present  results for other choices of $k_{S_i}$. 
We always let $C_0$ be a  randomly chosen subset of size $|C_0|=100$. 
The two heuristics perform surprisingly well.
 Although coming without any worst-case guarantees, the cost of their solutions is comparable to the cost of the output of our~algorithm.

\subsection{
Comparison with 
Unfair Algorithm~\ref{alg_greedy_standard}
}

We compare the cost of the solution produced by our  algorithm to the cost of the (potentially) unfair solution provided by Algorithm~\ref{alg_greedy_standard}. Of course, we expect the latter to be lower. We consider the case $k_{S_i}=k_{S_j}$, $i,j\in[m]$, and 
also 
examine how balanced the numbers of centers from a 
group~$S_i$ in the 
output of
Algorithm~\ref{alg_greedy_standard} are.  
Figure \ref{fig_comp_greedy} shows the results, where the data sets and settings equal the ones in the experiments 
of 
Figure~\ref{fig_comp_heuristics}. Similar experiments with different settings are provided in Figure~\ref{fig_comp_greedy_SUPPMAT} in Appendix~\ref{appendix_experiments}. 
Remarkably, the costs of the solutions produced by our algorithm and Algorithm~\ref{alg_greedy_standard} have the same order of magnitude in all experiments, showing that the price of fairness is 
small. 
On the other hand, 
the 
output of  Algorithm~\ref{alg_greedy_standard} can be highly~unfair.

\section{Discussion}\label{section_discussion}
In this work, 
we 
considered 
$k$-center clustering under a 
fairness constraint that is motivated by the application of centroid-based clustering for data summarization. 
We presented 
a simple 
approximation algorithm with running time only linear in the size of the data set $S$ and the number of centers~$k$ 
and 
proved our algorithm to be a 5-approximation algorithm when 
$S$ consists of two 
groups. 
For more than~two groups, 
we proved 
an upper bound on the approximation factor 
that  increases exponentially with the number of groups. 
We do not know whether this exponential dependence is 
necessary or whether 
our analysis is loose---in 
our extensive numerical simulations we \emph{never} observed a large approximation factor. 
Besides 
answering this question,  in future work it would be interesting to extend 
our results to 
$k$-medoid clustering 
or 
to characterize properties of data sets that guarantee that fast algorithms  find an optimal~fair~clustering.

\section*{Acknowledgements}

This research is supported by a Rutgers Research Council Grant and a Center for Discrete Mathematics and Theoretical Computer Science (DIMACS) postdoctoral fellowship.

\bibliography{mybibfile_fair_clustering}
\bibliographystyle{icml2019}


\clearpage

\icmltitlerunning{Appendix to Fair $k$-Center Clustering for Data Summarization}
\onecolumn
\appendix

\section*{Appendix}\label{appendix}

\section{Proofs}\label{appendix_proofs}

\vspace{1mm}
\textbf{Proof of Lemma \ref{lemma_greedy}:}

\vspace{2mm}
It is straightforward to see that Algorithm \ref{alg_greedy_standard} can be implemented in time $\mathcal{O}((k+|C_0'|) |S|)$. We only need to show that it is a 2-approximation algorithm for \eqref{unfair_k_center_general}.

If $k=0$, there is nothing to show, so  assume 
that 
$k\geq 1$. 
Let $C=\{c_1,\ldots,c_{k}\}$ be the output of Algorithm \ref{alg_greedy_standard} and $C^*=\{c_1^*,\ldots, c_k^*\}$ be an optimal solution to \eqref{unfair_k_center_general} with objective value~$r^*$. 
Let $s\in S$ be arbitrary. We need to show that $d(s,\hat{c})\leq 2r^*$ for some $\hat{c}\in C\cup C_0'$. 
If $s\in C\cup C_0'$, there is nothing to show. So assume $s\notin C\cup C_0'$. If
$$C_0'\cap \argmin_{c\in C^*\cup C_0'} d(s,c)\neq \emptyset,$$
there exists $\hat{c}\in C_0'$ with $d(s,\hat{c})\leq r^*$ and we are done. Otherwise, let $c_i^*\in \argmin_{c\in C^*\cup C_0'} d(s,c)$ and hence $d(s,c_i^*)\leq r^*$. We distinguish two cases:
\begin{itemize}
\item $\exists ~c_j \in C$ with $c_i^*\in \argmin_{c\in C^*\cup C_0'} d(c_j,c)$:

We have 
$d(c_j,c_i^*)\leq r^*$ and hence $d(s,c_j)\leq d(s,c_i^*)+d(c_i^*,c_j)\leq 2r^*$.

\item $\nexists ~c_j \in C$ with $c_i^*\in \argmin_{c\in C^*\cup C_0'} d(c_j,c)$:

There must be $c'\neq c''\in C\cup C_0'$, where not both $c'$ and $c''$ can be in $C_0'$, and $\hat{c}\in C^*\cup C_0'$ such that 
$$\hat{c}\in \argmin_{c\in C^*\cup C_0'} d(c',c)\cap\argmin_{c\in C^*\cup C_0'} d(c'',c).$$
Since $d(c',\hat{c})\leq r^*$ and $(c'',c^*)\leq r^*$, it follows that $d(c',c'')\leq d(c',\hat{c})+d(\hat{c},c'')\leq 2r^*$. 

\vspace{2mm}
Without loss of generality, assume that in the execution of Algorithm \ref{alg_greedy_standard}, $c''$ has been added to the set of centers after $c'$ has been added. In particular, we have $c''\in C$ and $c'' = c_l$ for some $l\in\{1,\ldots,k\}$. 
Due to the greedy choice in Line~5 of the algorithm and since $s$ has not been chosen by the algorithm, we have
\begin{align*}
2r^*\geq d(c',c'')\geq \min_{c\in \{c_1,\ldots,c_{l-1}\}\cup C_0'}d(c'',c)\geq \min_{c\in  \{c_1,\ldots,c_{l-1}\}\cup C_0'}d(s,c).
\end{align*}
\end{itemize}
 \hfill $\square$

\vspace{8mm}
\textbf{Proof of Theorem \ref{theorem_2groups}:}

\vspace{2mm}
Again it is easy to see that 
Algorithm~\ref{algorithm_2groups}  
can be implemented in time 
$\mathcal{O}((k+|C_0|) |S|)$.
We need to prove that it is a 5-approximation algorithm, but not a $(5-\varepsilon)$-approximation algorithm for any~$\varepsilon>0$:
\begin{enumerate}
 \item Algorithm \ref{algorithm_2groups} is a 5-approximation algorithm:
 
 \vspace{2mm}
Let 
$r^*_{\fair}$
be the optimal value of the fair 
problem \eqref{fair_k_center}
and
$r^*$ 
be the optimal value of the unfair
problem \eqref{unfair_k_center_general}. 
Clearly, $r^*\leq r^*_{\fair}$. 
Let  
$C^*_{\fair}=\{c^{(1)*}_1,\ldots,c^{(1)*}_{k_{S_1}},c^{(2)*}_1,\ldots,c^{(2)*}_{k_{S_2}}\}$ 
with $c^{(1)*}_1,\ldots,c^{(1)*}_{k_{S_1}}\in S_1$ and 
$c^{(2)*}_1,\ldots,c^{(2)*}_{k_{S_2}}\in S_2$
be an optimal solution to the  fair problem \eqref{fair_k_center} with cost~$r^*_{\fair}$ and 
$C^A=\{c_1^A,\ldots,c_k^A\}$ be the centers returned by Algorithm~\ref{algorithm_2groups}. 
It is clear that Algorithm \ref{algorithm_2groups} returns $k_{S_1}$ many elements from $S_1$ and 
$k_{S_2}$ many elements from $S_2$ and hence 
$C^A=\{c^{(1)A}_1,\ldots,c^{(1)A}_{k_{S_1}},c^{(2)A}_1,\ldots,c^{(2)A}_{k_{S_2}}\}$ with
$c^{(1)A}_1,\ldots,c^{(1)A}_{k_{S_1}}\in S_1$ and 
$c^{(2)A}_1,\ldots,c^{(2)A}_{k_{S_2}}\in S_2$.
We need to show that 
$$\min_{c\in C^A \cup C_0} d(s,c) \leq 5r^*_{\fair}, \quad s\in S.$$

\vspace{2mm}
Let $\widetilde{C}^A=\{\tilde{c}_1^A,\ldots,\tilde{c}_k^A\}$ be the output 
of Algorithm \ref{alg_greedy_standard} when called in Line 3 of Algorithm \ref{algorithm_2groups}. 
Since Algorithm~\ref{alg_greedy_standard} 
is a 2-approximation algorithm for the unfair problem \eqref{unfair_k_center_general} according to 
Lemma \ref{lemma_greedy}, 
we have 
\begin{align}\label{qw1}
\min_{c\in \widetilde{C}^A \cup C_0} d(s,c) \leq 2r^*\leq 2r^*_{\fair}, \quad s\in S. 
\end{align}
If Algorithm \ref{algorithm_2groups} returns $\widetilde{C}^A$ in Line 6, 
that is $C^A=\widetilde{C}^A$, we are done. Otherwise assume, as in the algorithm, 
that $|\widetilde{C}^A\cap S_1|>k_{S_1}$. Let $\tilde{c}_i^A \in S_1$ be a center of cluster $L_i$ 
that we replace with $y\in L_i\cap S_2$ and let $\hat{y}$ be an arbitrary element in $L_i$. Because 
of \eqref{qw1}, we have $d(\tilde{c}_i^A,y)\leq 2r^*_{\fair}$ and 
$d(\tilde{c}_i^A,\hat{y})\leq 2r^*_{\fair}$,  
and hence $d(y,\hat{y})\leq d(y,\tilde{c}_i^A)+d(\tilde{c}_i^A,\hat{y})\leq 4r^*_{\fair}$ due to the 
triangle inequality.  
Consequently, after the while-loop in Line 9, every $s\in S$ 
 is in distance of $4r^*_{\fair}$ or smaller to the center of its cluster. In particular, we have 
\begin{align*}
\min_{c\in \widetilde{C}^A \cup C_0} d(s,c) \leq 4r^*_{\fair}, \quad s\in S,
\end{align*}
and if Algorithm \ref{algorithm_2groups} returns $\widetilde{C}^A$ in Line 13, we are done.
Otherwise,  we still have 
$|\widetilde{C}^A\cap S_1|>k_{S_1}$ 
 after exchanging centers 
 in 
 the while-loop in Line~9. 
 Let 
 $S'=\cup_{i \in[k]: \tilde{c}_i^A\in S_1}L_i$, 
that is the union of clusters with a center $\tilde{c}_i^A\in S_1$. Since there is no more 
center in $S_1$ that we can exchange for an element in $S_2$, we have $S'\subseteq S_1$. Let 
$S''=\cup_{i \in[k]: \tilde{c}_i^A\in S_2}L_i$ 
be the union of clusters with a center $\tilde{c}_i^A\in S_2$ and $S_{C_0}=L_1'\cup \ldots\cup L_{|C_0|}'$ 
be the union of clusters with a center in $C_0$.
Then we have $S=S'\, \dot{\cup}\, S''\,\dot{\cup}\, S_{C_0}$.
We have $\widetilde{C}^A  \cap S_2 \subseteq C^A$ and  
\begin{align}\label{glA}
\min_{c\in C^A \cup C_0} d(s,c) \leq \min_{c\in(\widetilde{C}^A  \cap S_2) \cup C_0} d(s,c) \leq  4r^*_{\fair}, 
\quad s\in S''\cup S_{C_0}.
\end{align}
Hence we only need to show that 
$\min_{c\in C^A \cup C_0} d(s,c) \leq 5r^*_{\fair}$ for every $s\in S'$. 
We split $S'$ into two subsets $S'=S'_a\dot{\cup} S'_b$, where 
\begin{align*}
S'_a=\left\{s\in S': \argmin_{c\in  C^*_{\fair} \cup \, C_0} d(s,c) ~\cap~ (C_0 \cup S_2)\neq 
\emptyset \right\}
\end{align*}
and $S'_b=S'\setminus S'_a$. For every $s\in S'_a$ there is $c\in (C_0\cup S_2)\subseteq 
(S''\cup S_{C_0})$ with $d(s,c)\leq r^*_{\fair}$ and it follows from \eqref{glA} and 
the triangle inequality that 
\begin{align}\label{glB}
\min_{c\in C^A \cup C_0} d(s,c) \leq\min_{c\in(\widetilde{C}^A  \cap S_2) \cup C_0} d(s,c)\leq  5r^*_{\fair}, 
\quad s\in S'_a.
\end{align}
It remains to show that 
$\min_{c\in C^A \cup C_0} d(s,c) \leq 5r^*_{\fair}$ for every $s\in S'_b$.
For every $s\in S'_b$ there exists $c\in \{c_1^{(1)*},\ldots, c_{k_{S_1}}^{(1)*}\}$ with $d(s,c)\leq 
r^*_{\fair}$. 
We can write 
$S'_b=\cup_{j=1}^{k_{S_1}}\{s\in S'_b: d(s,c_j^{(1)*})\leq  r^*_{\fair}\}$ 
(some of the sets in this union might be empty, but that does not matter). 
Note that for every $j\in \{1,\ldots,k_{S_1}\}$ we have 
\begin{align}\label{glC}
d(s,s')\leq 2r^*_{\fair}, \quad s,s'\in \left\{s\in S'_b: d(s,c_j^{(1)*})\leq  r^*_{\fair}\right\}, 
\end{align}
due to the triangle inequality. 
It is 
\begin{align*}
S'=S'_a \cup S'_b=S'_a \cup \bigcup_{j=1}^{k_{S_1}}\left\{s\in S'_b: d(s,c_j^{(1)*})\leq  r^*_{\fair}\right\}
\end{align*}
 and when, in Line~15 of Algorithm~\ref{algorithm_2groups},  we run Algorithm~\ref{alg_greedy_standard}
 on $S' \cup C_0'$ with $k=k_{S_1}$ and initial centers $C_0'=C_0 \cup(\widetilde{C}^A  \cap S_2)$, 
 one of the following three cases has to happen (we denote the centers 
 returned by Algorithm~\ref{alg_greedy_standard} by $\widehat{C}^A=\{c^{(1)A}_1,\ldots,c^{(1)A}_{k_{S_1}}\}$):
 \begin{itemize}
 \item For every  $j\in\{1,\ldots,k_{S_1}\}$ there exists $j'\in\{1,\ldots,k_{S_1}\}$ such that 
 $c^{(1)A}_{j'}\in \{s\in S'_b: d(s,c_j^{(1)*})\leq  r^*_{\fair}\}$. In this case it immediately 
 follows from \eqref{glC} that
 \begin{align*}
\min_{c\in C^A \cup C_0} d(s,c) \leq \min_{c\in \widehat{C}^A} d(s,c)  \leq 2r^*_{\fair}, \quad s\in S'_b.
\end{align*}
\item There exists $j'\in\{1,\ldots,k_{S_1}\}$ such that $c^{(1)A}_{j'}\in S'_a$. 
When Algorithm~\ref{alg_greedy_standard} picks $c^{(1)A}_{j'}$, any other element in $S'$ cannot be at a larger
minimum distance from a center in $(\widetilde{C}^A  \cap S_2) \cup C_0$ or 
a previously 
chosen center in 
$\widehat{C}^A$ than $c^{(1)A}_{j'}$. It follows from \eqref{glB} that
\begin{align*}
\min_{c\in C^A \cup C_0} d(s,c) \leq 5r^*_{\fair}, \quad s\in S'.
\end{align*}
\item There exist  $j\in\{1,\ldots,k_{S_1}\}$ and  $j'\neq j''\in\{1,\ldots,k_{S_1}\}$ such that 
$c^{(1)A}_{j'},c^{(1)A}_{j''}\in \{s\in S'_b: d(s,c_j^{(1)*})\leq  r^*_{\fair}\}$. Assume that 
Algorithm~\ref{alg_greedy_standard}
picks $c^{(1)A}_{j'}$ before $c^{(1)A}_{j''}$. When Algorithm~\ref{alg_greedy_standard} picks $c^{(1)A}_{j''}$, 
any other element in $S'$ cannot be at a larger
minimum distance from a center in $(\widetilde{C}^A  \cap S_2) \cup C_0$ or 
a previously 
chosen center in 
$\widehat{C}^A$ than $c^{(1)A}_{j''}$.  
Because of $d(c^{(1)A}_{j'},c^{(1)A}_{j''})\leq 2r^*_{\fair}$ according to \eqref{glC}, it follows that 
\begin{align*}
\min_{c\in C^A \cup C_0} d(s,c) \leq 2r^*_{\fair}, \quad s\in S'.
\end{align*}
 \end{itemize}
 In all cases we have 
 $$\min_{c\in C^A \cup C_0} d(s,c) \leq 5r^*_{\fair}, \quad s\in S'_b,$$
 which completes the proof of the claim that Algorithm \ref{algorithm_2groups} is a 5-approximation algorithm.
 
 \item Algorithm \ref{algorithm_2groups} is not a $(5-\varepsilon)$-approximation algorithm for any~$\varepsilon>0$:

 \begin{figure}[t]
 \centering
\includegraphics[scale=0.9]{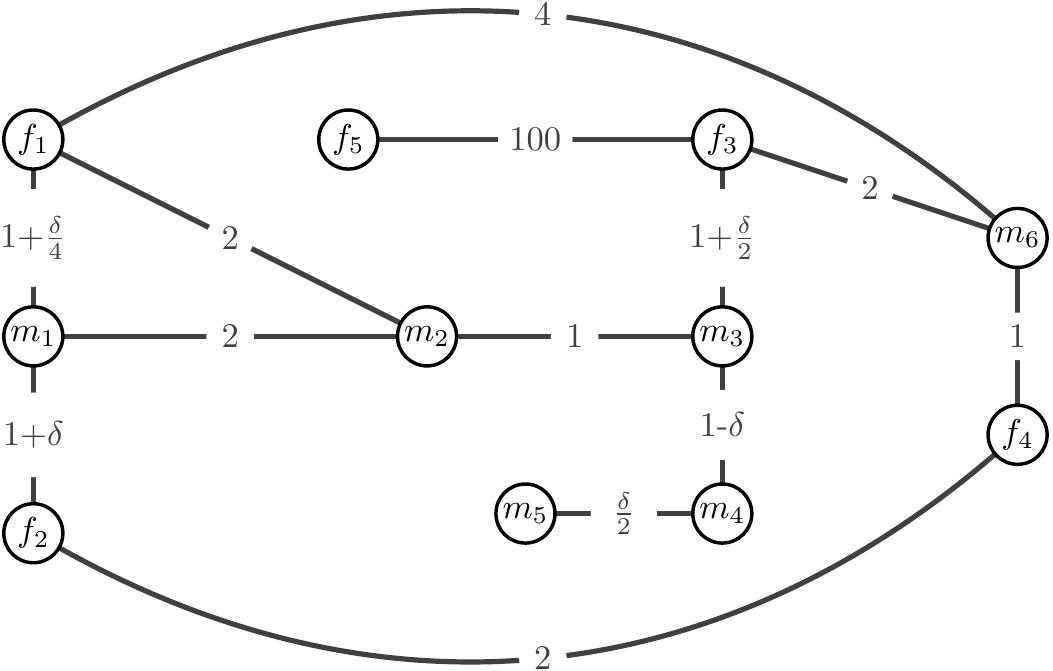}
\caption{An example showing that Algorithm \ref{algorithm_2groups} is 
not a $(5-\varepsilon)$-approximation algorithm for any~$\varepsilon>0$.}\label{sketch_lower_bound_two_groups}
 \end{figure}

 \vspace{2mm} 
 Consider the 
 example 
 given by 
 the 
 weighted 
 graph shown in Figure \ref{sketch_lower_bound_two_groups}, 
 where $0<\delta<\frac{1}{10}$. 
We have 
$S=S_1\dot{\cup} S_2$ 
with  $S_1=\{f_1,f_2,f_3,f_4,f_5\}$ and $S_2=\{m_1,m_2,m_3,m_4,m_5,m_6\}$. 
All distances are shortest-path-distances. Let $k_{S_1}=1$, $k_{S_2}=3$, and $C_0=\emptyset$. 
We assume that Algorithm~\ref{alg_greedy_standard} in 
Line~3 of Algorithm~\ref{algorithm_2groups} picks $f_5$ as first center. It then chooses  $f_2$ as second center, 
$f_3$ as 
third center and $f_1$ as fourth center. Hence,  $\widetilde{C}^A=\{f_5,f_2,f_3,f_1\}$ and 
 $|\widetilde{C}^A\cap S_1|>k_{S_1}$. 
The clusters corresponding  to $\widetilde{C}^A$ are $\{f_5\}$, $\{f_2,f_4\}$,  
$\{f_3,m_3,m_4,m_5,m_6\}$ and $\{f_1,m_1,m_2\}$. Assume we replace $f_3$ with $m_4$ and  
$f_1$ with $m_2$ in Line~10 of Algorithm~\ref{algorithm_2groups}. Then it is still 
$|\widetilde{C}^A\cap S_1|>k_{S_1}$, and 
in Line~15 of Algorithm~\ref{algorithm_2groups} we run Algorithm~\ref{alg_greedy_standard} on 
$\{f_2,f_4,f_5\}\cup\{m_2,m_4\}$ with $k=1$ and initially given centers $C_0'=\{m_2,m_4\}$. 
Algorithm~\ref{alg_greedy_standard} returns $\widehat{C}^A=\{f_5\}$. Finally, assume that $m_5$ is chosen as 
arbitrary third center from $S_2$ in Line~16 of Algorithm~\ref{algorithm_2groups}. 
So the centers returned by Algorithm~\ref{algorithm_2groups}
are $C^A=\{f_5,m_2,m_4,m_5\}$ with a cost of $5-\frac{\delta}{2}$ (incurred for $f_4$). However, 
the optimal solution  $C^*_{\fair}=\{f_5,m_1,m_3,m_6\}$ has cost only $1+\delta$.
Choosing $\delta$ sufficiently small shows  that Algorithm \ref{algorithm_2groups} is 
not a $(5-\varepsilon)$-approximation algorithm for any~$\varepsilon>0$.
 \end{enumerate}
\hfill $\square$

\vspace{8mm}
\textbf{Proof of Lemma \ref{lemma_alg_helper}:}

\vspace{2mm}
We want to show three things:

\begin{enumerate}
  \item 
Algorithm \ref{alg_helper} is well-defined:  

If the condition of the while-loop in Line 7 is true, 
there exists a 
shortest path $P=S_{v_0}S_{v_1}\cdots S_{v_w}$ with $S_{v_0}=S_r$, $S_{v_w}=S_s$ that connects $S_r$ to $S_s$ in $G$. 
Since $P$ is a shortest path, all $S_{v_i}$ are distinct. By the definition of $G$, for every $l=0,\ldots,w-1$ there 
exists $L_t$ with center $\tilde{c}_t^A\in S_{v_l}$ and $y\in L_t\cap S_{v_{l+1}}$. 
Hence, the for-loop in Line 8 is well defined.

\item 
Algorithm \ref{alg_helper} terminates: 

Let, at the beginning of the execution of Algorithm 
\ref{alg_helper} in Line 3, $H_1=\{S_j\in\{S_1,\ldots,S_m\}: \tilde{k}_{S_j}= k_{S_j}\}$, 
$H_2=\{S_j\in\{S_1,\ldots,S_m\}: \tilde{k}_{S_j}> k_{S_j}\}$ 
and 
$H_3=\{S_j\in\{S_1,\ldots,S_m\}: \tilde{k}_{S_j}< k_{S_j}\}$. For $S_j\in H_1$, $\tilde{k}_{S_j}$ never 
changes during the 
execution of the algorithm. For $S_j\in H_2$, $\tilde{k}_{S_j}$ never increases during 
the execution of the algorithm and decreases 
at most until it equals $k_{S_j}$. For $S_j\in H_3$, 
$\tilde{k}_{S_j}$ never decreases during the execution of the algorithm and increases at most 
until it equals $k_{S_j}$. 
In every iteration of the while-loop, 
there is 
$S_j\in H_3$ for which $\tilde{k}_{S_j}$ increases by one. 
It follows that the number of iterations of the 
while-loop is upper-bounded by $k$.

\item 
Algorithm \ref{alg_helper} exchanges centers in such a way that the set~$\mathcal{G}$ that it returns satisfies 
$\mathcal{G}\subsetneq \{S_1,\ldots,S_m\}$
and properties~\eqref{property_G_1} and \eqref{property_G_2}: 

Note that throughout the execution of Algorithm \ref{alg_helper} we have 
$\tilde{k}_{S_j}=\sum_{i=1}^k \charfct\left\{\tilde{c}_i^A\in S_j\right\}$ for the current 
centers $\tilde{c}_1^A,\ldots,\tilde{c}_k^A$.
If the condition of the if-statement in Line 13 is true, 
then $\mathcal{G}=\emptyset$ and \eqref{property_G_1} and \eqref{property_G_2} are satisfied. 

Assume that the condition  of the if-statement in Line 13 is not true. 
Clearly, the set $\mathcal{G}$ returned by 
Algorithm~\ref{alg_helper} satisfies~\eqref{property_G_2}. Since the condition 
of the if-statement in Line 13 is not true, there exist $S_{j}$ with $\tilde{k}_{S_j}> k_{S_j}$ and $S_i$ 
with  $\tilde{k}_{S_i}< k_{S_i}$. 
We have $S_j\in \mathcal{G}$, but since the condition of the  while-loop in Line 7 is not true, we 
cannot have $S_i\in \mathcal{G}$. This shows 
that $\mathcal{G}\subsetneq \{S_1,\ldots,S_m\}$. We need to show that \eqref{property_G_1} holds. Let $L_h$ 
be a cluster with center  $\tilde{c}_h^A\in S_f$ for some 
$S_f \in\mathcal{G}$ and assume 
it contained an element $o\in S_{f'}$ with $S_{f'} \notin \mathcal{G}$. But then we had a path from 
$S_f$ to $S_{f'}$ in $G$. 
If $S_f\in \mathcal{G}'$, 
this is an immediate contradiction to $S_{f'} \notin \mathcal{G}$. 
If $S_f\notin \mathcal{G}'$, since $S_f \in\mathcal{G}$, there exists $S_g\in \mathcal{G}'$ 
such that there is a path from 
$S_g$ to $S_f$. But then there is also a path from $S_g$ to $S_{f'}$, which is a contradiction 
to $S_{f'} \notin \mathcal{G}$.
\end{enumerate}
\hfill $\square$

\vspace{8mm}
\textbf{Proof of Theorem \ref{theorem_mgroups}:}

\vspace{2mm}

For showing that Algorithm \ref{algorithm_mgroups} is a $(3\cdot 2^{m-1}-1)$-approximation algorithm let 
$r^*_{\fair}$
be the optimal value of problem \eqref{fair_k_center} and 
$C^*_{\fair}$
be an optimal solution with cost~$r^*_{\fair}$.  Let  
$C^A$ be the centers returned by Algorithm \ref{algorithm_mgroups}. A simple proof by induction over~$m$ 
shows that $C^A$ actually comprises $k_{S_i}$ many elements from every group $S_i$. 
We need to show that 
\begin{align}\label{to_show_theorem2}
\min_{c\in C^A \cup C_0} d(s,c) \leq (3\cdot 2^{m-1}-1)r^*_{\fair}, \quad s\in S.
\end{align}

Let $T$ be the total number of calls of Algorithm~\ref{algorithm_mgroups}, that is we have one initial call and 
$T-1$ recursive calls. 
Since with each recursive call the number of groups is decreased by at least one,   
we have $T\leq m$. 
For $1\leq j \leq T$, let $S^{(j)}$ be the data set in the $j$-th call of Algorithm \ref{algorithm_mgroups}.
We additionally set $S^{(T+1)}=\emptyset$. We have  
$S^{(1)}=S$ and $S^{(j)}\supseteq S^{(j+1)}$, $1\leq j \leq T$.
For $1\leq j <T$, let $\mathcal{G}^{(j)}$ be the set of groups in $\mathcal{G}$ 
returned by Algorithm \ref{alg_helper} in Line~8 in the $j$-th call of Algorithm~\ref{algorithm_mgroups}. 
If in the $T$-th call of Algorithm \ref{algorithm_mgroups} the algorithm terminates from Line~10 
(note that in this case we must have $T<m$), we also let
$\mathcal{G}^{(T)}=\emptyset$ be the set of groups  
in $\mathcal{G}$ 
returned by Algorithm \ref{alg_helper} 
in the $T$-th call. Otherwise we leave $\mathcal{G}^{(T)}$ undefined.
Setting 
$\mathcal{G}^{(0)}=\{S_1,\ldots,S_m\}$,  
we have $\mathcal{G}^{(j)}\supsetneq \mathcal{G}^{(j+1)}$ for all $j$ such that $\mathcal{G}^{(j+1)}$ is defined. 
For $1\leq j <T$, let $C_j$ be the set of centers 
returned by Algorithm~\ref{alg_helper} in Line~8 in the $j$-th call of Algorithm~\ref{algorithm_mgroups} that 
belong to a group not in $\mathcal{G}^{(j)}$ (in Algorithm~\ref{algorithm_mgroups}, 
the set of these centers is denoted by $C'$). 
We analogously define  $C_T$ if in the $T$-th call of Algorithm~\ref{algorithm_mgroups} 
the algorithm terminates from Line 10. 
Note that the centers in $C_j$ are comprised in the final output $C^A$ of Algorithm~\ref{algorithm_mgroups}, 
that is $C_j\subseteq C^A$ for $1\leq j <T$ or $1\leq j \leq T$. 
As always, 
$C_0$ denotes the set of centers that are 
given initially (for the initial call of  Algorithm~\ref{algorithm_mgroups}). Note that in the $j$-th call of 
Algorithm~\ref{algorithm_mgroups} the set of initially given centers is $C_0 \cup \bigcup_{l=1}^{j-1} C_l$.

\vspace{5mm}
We first prove by induction that for all $j\geq 1$ such that $\mathcal{G}^{(j)}$ is defined, 
that is $1\leq j <T$ or $1\leq j \leq T$, we have
\begin{align}\label{prop_ind}
\min_{c\in  C_0 \cup \bigcup_{l=1}^j C_l} d(s,c)\leq (2^{j+1}+2^j-2)r^*_{\fair}, \quad 
s\in \left(S^{(j)} \setminus S^{(j+1)}\right)\cup  \left(C_0 \cup \bigcup_{l=1}^j C_l\right).
\end{align}

\vspace{2mm}
\textbf{Base case} $j=1$: In the first call of Algorithm \ref{algorithm_mgroups}, 
Algorithm \ref{alg_greedy_standard}, when called in Line 3 of Algorithm \ref{algorithm_mgroups}, returns an 
approximate solution to the unfair problem \eqref{unfair_k_center_general}. 
Let $r^*\leq r^*_{\fair}$ be the optimal cost of \eqref{unfair_k_center_general}. 
Since Algorithm \ref{alg_greedy_standard} is a 2-approximation algorithm for  \eqref{unfair_k_center_general} 
according to Lemma \ref{lemma_greedy}, 
after Line 3 of Algorithm \ref{algorithm_mgroups} we have 
\begin{align*}
\min_{c\in \widetilde{C}^A \cup C_0} d(s,c) \leq 2r^*\leq 2r^*_{\fair}, \quad s\in S.
\end{align*}
Let $\tilde{c}_i^A \in \widetilde{C}^A$ be a center and $s_1,s_2\in L_i$ be two points in its cluster. 
It follows from the triangle inequality that 
$d(s_1,s_2)\leq d(s_1,\tilde{c}_i^A)+d(\tilde{c}_i^A,s_2)\leq 4r^*_{\fair}$. Hence, after running 
Algorithm~\ref{alg_helper} in Line 8 of Algorithm \ref{algorithm_mgroups} and exchanging 
some of the centers in $\widetilde{C}^A$, we have 
$d(s,c(s))\leq 4r^*_{\fair}$ for every $s\in S$, where $c(s)$ denotes the center of its cluster.
In particular, 
\begin{align*}
\min_{c\in C_0 \cup C_1} d(s,c)\leq (2^{1+1}+2^1-2)r^*_{\fair}=4r^*_{\fair}
\end{align*}
for all $s\in S$ for which its center $c(s)$ is in $C_0$ or in a group not in $\mathcal{G}^{(1)}$, that is for 
$s\in (S^{(1)} \setminus S^{(2)})\cup  (C_0 \cup C_1)$.

\vspace{2mm}
\textbf{Inductive step} $j\mapsto j+1$: Recall property \eqref{property_G_1} of a set $\mathcal{G}$
returned by Algorithm~\ref{alg_helper}. Consequently, 
$S^{(j+1)}$ only comprises items in a group in 
$\mathcal{G}^{(j)}$ and, additionally, the given centers $C_0 \cup \bigcup_{l=1}^j C_l$.

We split $S^{(j+1)}$ into two subsets $S^{(j+1)}=S^{(j+1)}_a\dot{\cup} S^{(j+1)}_b$, where 
\begin{align*}
S^{(j+1)}_a=\left\{s\in S^{(j+1)}: \argmin_{c\in  C^*_{\fair} \cup C_0} d(s,c) ~\cap~ 
\left(C_0\cup \bigcup_{W\in \{S_1,\ldots,S_m\}\setminus \mathcal{G}^{(j)}} W\right)\neq \emptyset \right\}
\end{align*}
and $S^{(j+1)}_b=S^{(j+1)}\setminus S^{(j+1)}_a$. For every $s\in S^{(j+1)}_a$ there exists 
\begin{align*}
c\in C_0\cup \bigcup_{W\in \{S_1,\ldots,S_m\}\setminus \mathcal{G}^{(j)}} W\subseteq 
\left(S\setminus S^{(j+1)}\right)\cup \left(C_0 \cup \bigcup_{l=1}^j C_l\right)
\end{align*}
with $d(s,c)\leq r^*_{\fair}$. It follows from the inductive hypothesis that there 
exists $c'\in C_0 \cup \bigcup_{l=1}^j C_l$ 
with $d(c,c')\leq (2^{j+1}+2^j-2)r^*_{\fair}$ and consequently
\begin{align*}
d(s,c')\leq d(s,c)+d(c,c')\leq r^*_{\fair}+(2^{j+1}+2^j-2)r^*_{\fair}=(2^{j+1}+2^j-1)r^*_{\fair}.
\end{align*}
Hence,
\begin{align}\label{pr_qu}
\min_{c\in  C_0 \cup \bigcup_{l=1}^j C_l} d(s,c)\leq (2^{j+1}+2^j-1)r^*_{\fair},\quad s\in S^{(j+1)}_a.
\end{align}
For every $s\in S^{(j+1)}_b$ there exists $c\in C^*_{\fair}\cap \bigcup_{W\in \mathcal{G}^{(j)}} W$ with 
$d(s,c)\leq  r^*_{\fair}$. 
Let $C^*_{\fair}\cap \bigcup_{W\in \mathcal{G}^{(j)}} W=\{\tilde{c}^*_1,\ldots,\tilde{c}^*_{\tilde{k}}\}$ with 
$\tilde{k}=\sum_{W\in \mathcal{G}^{(j)}}k_W$, where $k_W$ is the number of requested centers from group $W$.  
We can write 
\begin{align*}
S^{(j+1)}_b=\bigcup_{l=1}^{\tilde{k}} \left\{s\in S^{(j+1)}_b: d(s,\tilde{c}^*_l)\leq  r^*_{\fair}\right\},
\end{align*}
where some of the sets in this union might be empty, but that does not matter. Note that for 
every $l=1,\ldots,\tilde{k}$ we have
\begin{align}\label{absch_diam_casem}
d(s,s')\leq 2r^*_{\fair}, \quad s,s'\in \left\{s\in S^{(j+1)}_b: d(s,\tilde{c}^*_l)\leq  r^*_{\fair}\right\}
\end{align}
due to the triangle inequality. It is 
\begin{align*}
S^{(j+1)}=S^{(j+1)}_a\cup S^{(j+1)}_b=S^{(j+1)}_a\cup
\bigcup_{l=1}^{\tilde{k}} \left\{s\in S^{(j+1)}_b: d(s,\tilde{c}^*_l)\leq  r^*_{\fair}\right\}
\end{align*}
 and when, in Line 3 of Algorithm \ref{algorithm_mgroups},  we run Algorithm \ref{alg_greedy_standard} 
 on $S^{(j+1)}$ with $k=\tilde{k}$ and initial centers 
 $C_0 \cup \bigcup_{l=1}^j C_l$,
 one of the following three cases has to happen 
 (we denote the centers returned by Algorithm \ref{alg_greedy_standard} in this $(j+1)$-th call 
 of Algorithm \ref{algorithm_mgroups} by 
 $\widetilde{F}^A=\{\tilde{f}_1^A,\ldots, \tilde{f}_{\tilde{k}}^A\}$ and assume that for 
 $1\leq l<l'\leq \tilde{k}$ 
 Algorithm \ref{alg_greedy_standard}
  has chosen  $\tilde{f}_l^A$  before~$\tilde{f}_{l'}^A$):
 \begin{itemize}
 \item For every  $l\in\{1,\ldots,\tilde{k}\}$ there exists $l'\in\{1,\ldots,\tilde{k}\}$ such that 
 $\tilde{f}_{l'}^A\in \{s\in S^{(j+1)}_b: d(s,\tilde{c}^*_l)\leq  r^*_{\fair}\}$. In this case it 
 immediately follows that
 \begin{align*}
\min_{c\in \widetilde{F}^A} d(s,c) \leq 2r^*_{\fair}, \quad s\in S^{(j+1)}_b, 
\end{align*}
and using \eqref{pr_qu} we obtain
\begin{align*} 
\min_{c\in C_0 \cup \bigcup_{l=1}^j C_l\cup \widetilde{F}^A} d(s,c) \leq (2^{j+1}+2^j-1)r^*_{\fair},\quad s\in S^{(j+1)}.
\end{align*}
\item There exists $l'\in\{1,\ldots,\tilde{k}\}$ such that $\tilde{f}_{l'}^A\in S^{(j+1)}_a$. 
When Algorithm \ref{alg_greedy_standard} picks $\tilde{f}_{l'}^A$, 
any other element in $S^{(j+1)}$ cannot be at a larger minimum distance from a center in $C_0 \cup \bigcup_{l=1}^j C_l$ or 
an already chosen center in $\{\tilde{f}_{l'}^A,\ldots,\tilde{f}_{l'-1}^A\}$ than $\tilde{f}_{l'}^A$. It follows from \eqref{pr_qu} that
\begin{align*} 
\min_{c\in C_0 \cup \bigcup_{l=1}^j C_l\cup \widetilde{F}^A} d(s,c) \leq (2^{j+1}+2^j-1)r^*_{\fair},\quad s\in S^{(j+1)}.
\end{align*}
\item There exist $l\in\{1,\ldots,\tilde{k}\}$ and  $l',l''\in\{1,\ldots,\tilde{k}\}$ with $l'<l''$ such that 
$\tilde{f}_{l'}^A,\tilde{f}_{l''}^A\in \{s\in S^{(j+1)}_b: d(s,\tilde{c}^*_{l})\leq  r^*_{\fair}\}$. 
When Algorithm \ref{alg_greedy_standard} picks $\tilde{f}_{l''}^A$, 
any other element in $S^{(j+1)}$ cannot be at a larger minimum distance from a center in 
$C_0 \cup \bigcup_{l=1}^j C_l$ or 
an already chosen center in $\{\tilde{f}_{l'}^A,\ldots,\tilde{f}_{l''-1}^A\}$ than $\tilde{f}_{l''}^A$. 
Because of $d(\tilde{f}_{l'}^A,\tilde{f}_{l''}^A)\leq 2r^*_{\fair}$ 
according to \eqref{absch_diam_casem},
it follows that 
\begin{align*} 
\min_{c\in C_0 \cup \bigcup_{l=1}^j C_l\cup \widetilde{F}^A} d(s,c)\leq 2r^*_{\fair} \leq (2^{j+1}+2^j-1)r^*_{\fair},\quad s\in S^{(j+1)}.
\end{align*}
\end{itemize}
In any case, we have
\begin{align}\label{central_eq_indstep} 
\min_{c\in C_0 \cup \bigcup_{l=1}^j C_l\cup \widetilde{F}^A} d(s,c) \leq (2^{j+1}+2^j-1)r^*_{\fair},\quad s\in S^{(j+1)}.
\end{align}
Similarly to the base case, it follows from the triangle inequality
that after running 
Algorithm~\ref{alg_helper} in Line 8 of Algorithm \ref{algorithm_mgroups} and exchanging 
some of the centers in $\widetilde{F}^A$, we have 
\begin{align*}
d(s,c(s))\leq 2(2^{j+1}+2^j-1)r^*_{\fair}=(2^{j+2}+2^{j+1}-2)r^*_{\fair}
\end{align*}
for every $s\in S^{(j+1)}$, where $c(s)$ denotes the center of its cluster. In particular,
we have
\begin{align*}
\min_{c\in C_0 \cup \bigcup_{l=1}^{j+1} C_l} d(s,c)\leq (2^{j+2}+2^{j+1}-2)r^*_{\fair}, 
\quad s\in \left(S^{(j+1)} \setminus S^{(j+2)}\right)\cup  \left( C_0 \cup \bigcup_{l=1}^{j+1} C_l\right),
\end{align*}
 and this completes the proof of \eqref{prop_ind}.

\vspace{5mm}
If in the $T$-th call of Algorithm \ref{algorithm_mgroups} the algorithm terminates from Line~10, 
it follows from \eqref{prop_ind} that
\begin{align}\label{gl_www}
 \min_{c\in C_0 \cup \bigcup_{l=1}^T C_l} d(s,c)\leq (2^{T+1}+2^T-2)r^*_{\fair}, \quad s\in S.
\end{align}
In this case, since $T<m$, we have 
\begin{align*}
 2^{T+1}+2^T-2\leq 2^m+2^{m-1}-2<2^m+2^{m-1}-1,
\end{align*}
and \eqref{gl_www} implies \eqref{to_show_theorem2}.
If in the $T$-th call of Algorithm~\ref{algorithm_mgroups} the algorithm does not terminate from Line~10, 
it must terminate from Line~5. 
It follows from \eqref{prop_ind} that
\begin{align}\label{gl_www2}
 \min_{c\in C_0\cup \bigcup_{l=1}^{T-1} C_l} d(s,c)\leq (2^{T}+2^{T-1}-2)r^*_{\fair}, \quad 
 s\in \left(S\setminus S^{(T)}\right)\cup \left(C_0 \cup \bigcup_{l=1}^{T-1} C_l\right).
\end{align}
In the same way as we have shown \eqref{central_eq_indstep} in the inductive step in the proof of \eqref{prop_ind},
we can show that 
\begin{align}\label{gl_www22} 
\min_{c\in C_0 \cup \bigcup_{l=1}^{T-1} C_l\cup \widetilde{H}^A} d(s,c)\leq (2^{T}+2^{T-1}-1)r^*_{\fair}\leq (2^{m}+2^{m-1}-1)r^*_{\fair},
\quad s\in S^{(T)},
\end{align}
where $\widetilde{H}^A$ is the set of centers returned by Algorithm \ref{alg_greedy_standard}
in the $T$-th call of Algorithm \ref{algorithm_mgroups}. Since $\bigcup_{l=1}^{T-1} C_l\cup \widetilde{H}^A$ 
is contained in the output $C^A$ of Algorithm~\ref{algorithm_mgroups}, \eqref{gl_www22} 
together with 
 \eqref{gl_www2}
implies \eqref{to_show_theorem2}.

\vspace{2mm}
Since running Algorithm \ref{algorithm_mgroups} involves at most $m$ (recursive) calls 
of the algorithm 
and the running time of each of these calls is dominated by the running times of 
Algorithm~\ref{alg_greedy_standard} and Algorithm~\ref{alg_helper}, it follows that the running 
time of  Algorithm \ref{algorithm_mgroups} is 
 $\mathcal{O}((|C_0|m+k m^2) |S|+km^4)$.
\hfill $\square$


\begin{figure}[t]
 \centering
\includegraphics[scale=0.9]{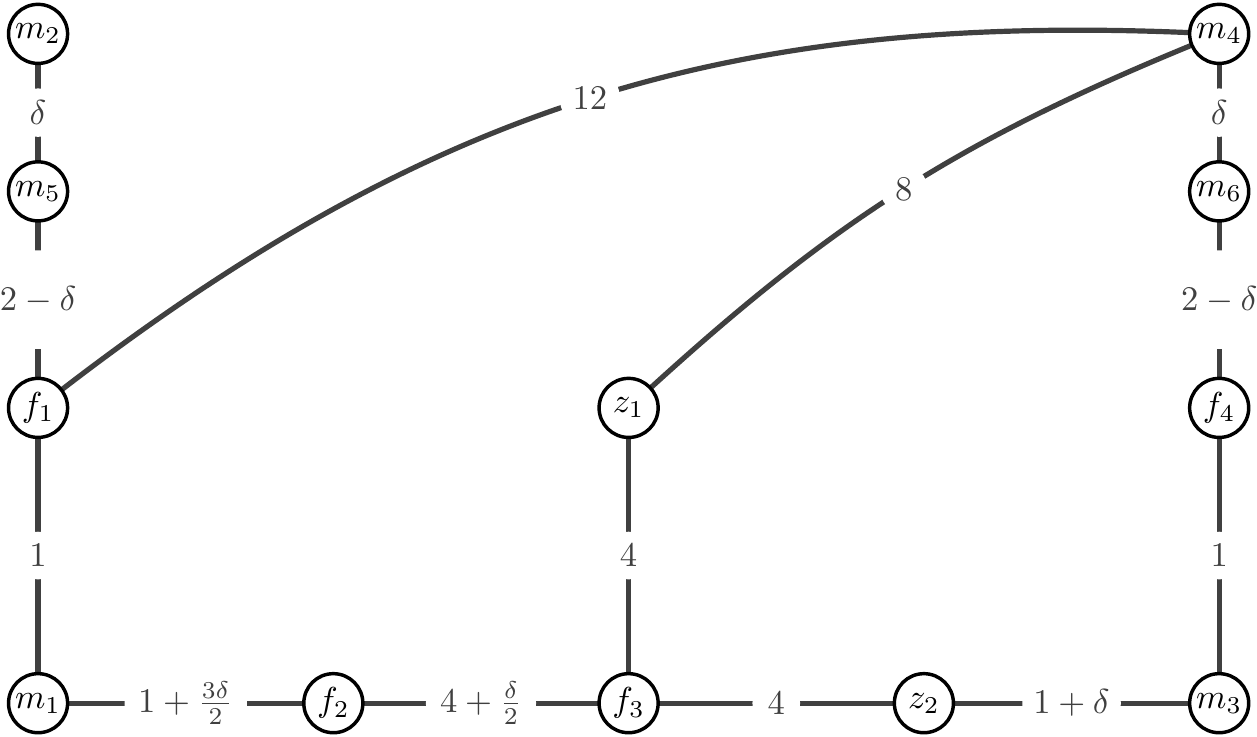}
\caption{An example showing that Algorithm~\ref{algorithm_mgroups} is 
not a $(8-\varepsilon)$-approximation algorithm for any~$\varepsilon>0$.}\label{sketch_lower_bound_3_groups}
 \end{figure}

\vspace{8mm}
\textbf{Proof of Lemma \ref{mgroups_lower_bound}:}

 \vspace{2mm} 
 Consider the 
 example 
 given by 
 the 
 weighted 
 graph shown in Figure \ref{sketch_lower_bound_3_groups}, 
 where $0<\delta<\frac{1}{10}$. 
We have 
$S=S_1\dot{\cup} S_2\dot{\cup} S_3$ 
with  $S_1=\{m_1,m_2,m_3,m_4,m_5,m_6\}$, $S_2=\{f_1,f_2,f_3,f_4\}$ and $S_3=\{z_1,z_2\}$. 
All distances are shortest-path-distances. Let $k_{S_1}=4$, $k_{S_2}=1$, $k_{S_3}=1$ and $C_0=\emptyset$. 
We assume that Algorithm~\ref{alg_greedy_standard} in 
Line~3 of Algorithm~\ref{algorithm_mgroups} picks $f_1$ as first center.
 It then chooses  $f_4$ as second center, $z_1$ as third center,  $f_3$ as fourth center,  
$f_2$ as fifth center and $z_2$ as sixth center. 
Hence,  $\widetilde{C}^A=\{f_1,f_4,z_1,f_3,f_2,z_2\}$ and 
the corresponding clusters are 
$\{f_1,m_1,m_2, m_5\}$, $\{f_4,m_3,m_4,m_6\}$, $\{z_1\}$, $\{f_3\}$, $\{f_2\}$ and $\{z_2\}$. 
When running Algorithm \ref{alg_helper} 
in Line~8 of Algorithm~\ref{algorithm_mgroups}, 
it replaces $f_1$ with one of $m_1$, $m_2$ or $m_5$ and it replaces $f_4$ with one of $m_3$, $m_4$ or $m_6$. 
Assume that it replaces $f_1$ with $m_2$ and $f_4$ with $m_4$. Algorithm \ref{alg_helper} 
then returns $\mathcal{G}=\{S_2,S_3\}$ and when recursively 
calling Algorithm~\ref{algorithm_mgroups} in Line 12, 
 we have $S'=\{f_2,f_3,z_1,z_2\}$ and $C'=\{m_2,m_4\}$. 
In the recursive call, the given centers are $C'$ and
Algorithm~\ref{alg_greedy_standard} chooses $f_3$ and $f_2$. 
The corresponding clusters are  $\{f_3,z_1,z_2\}$, $\{f_2\}$, $\{m_2\}$ and $\{m_4\}$.  
When running Algorithm~\ref{alg_helper} with clusters $\{f_3,z_1,z_2\}$ and $\{f_2\}$,
it replaces $f_3$ with either $z_1$ or $z_2$ and returns $\mathcal{G}=\emptyset$, that is afterwards we are done. 
Assume Algorithm \ref{alg_helper} replaces $f_3$ with $z_2$. 
Then the centers returned by Algorithm~\ref{algorithm_mgroups} are $z_2,f_2,m_2,m_4$ and two 
arbitrary elements from $S_1$, which we assume to be $m_5$ and $m_6$. These centers have
a cost of $8$ (incurred for $z_1$). 
However, 
an optimal solution such as $C^*_{\fair}=\{m_1,m_2,m_3,m_4,f_3,z_1\}$ has cost only $1+\frac{3\delta}{2}$.
Choosing $\delta$ sufficiently small shows  that Algorithm \ref{algorithm_mgroups} is 
not a $(8-\varepsilon)$-approximation algorithm for any~$\varepsilon>0$.
\hfill $\square$

\vspace{10mm}
\section{Further Experiments}\label{appendix_experiments}

\begin{figure*}[h]
\centering
\includegraphics[width=0.88\columnwidth]{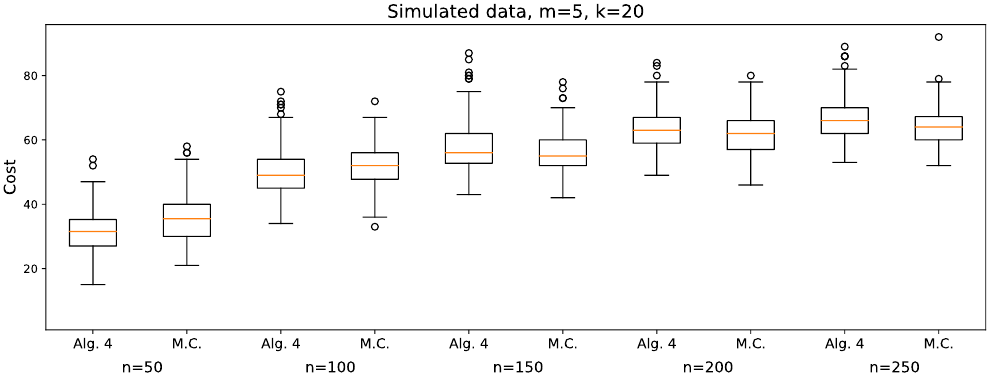}
\caption{Cost of the output of our algorithm (Alg.~\ref{algorithm_mgroups}) in comparison to the algorithm by \citeauthor{chen_matroid_center} (M.C.) in the run-time experiment shown in the right part of Figure~\ref{fig_exp_comparison_matroid_k_center}.}\label{fig_exp_cost_comp_matr_intersection}
\end{figure*}

\begin{figure}[h]
\centering
\includegraphics[scale=0.35]{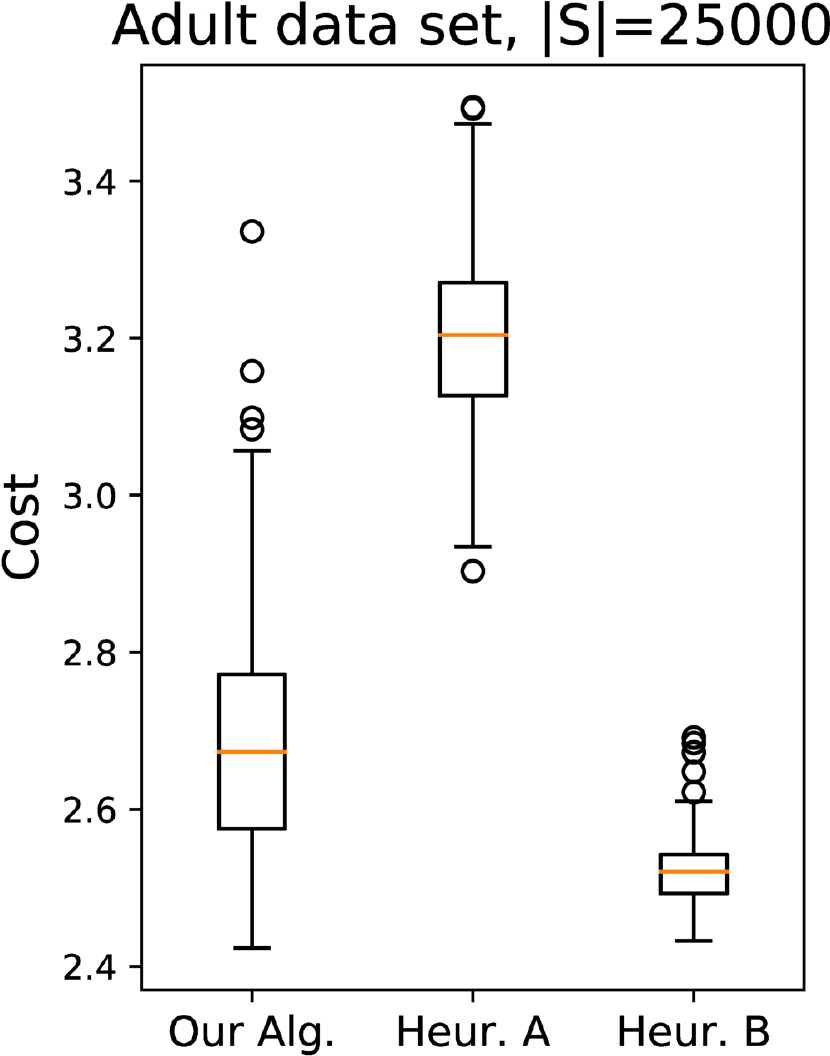}
\hspace{6mm}
\includegraphics[scale=0.35]{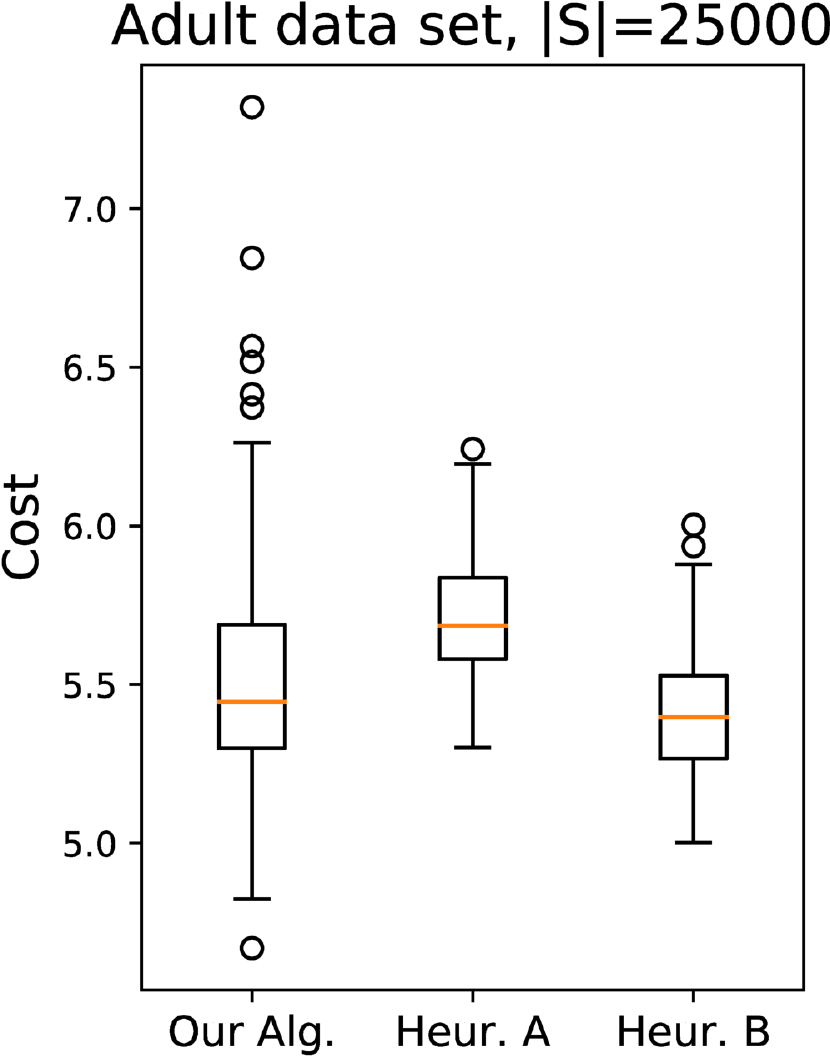}
\hspace{6mm}
\includegraphics[scale=0.35]{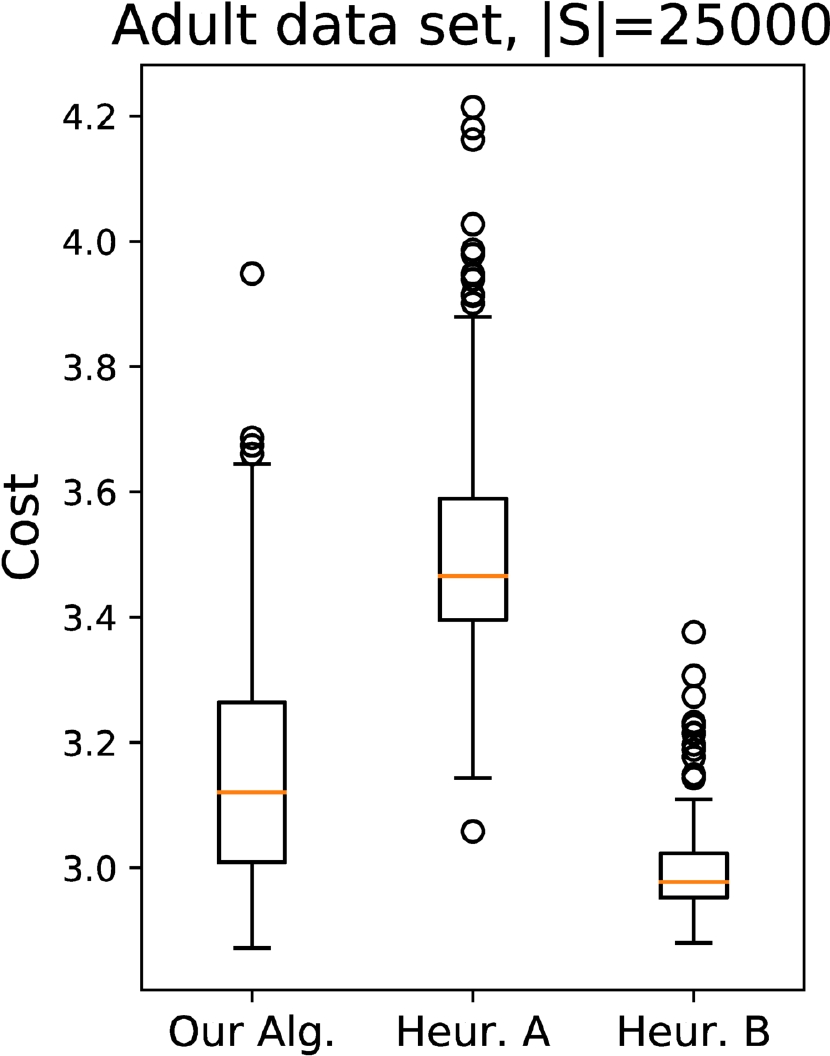}
\hspace{6mm}
\includegraphics[scale=0.35]{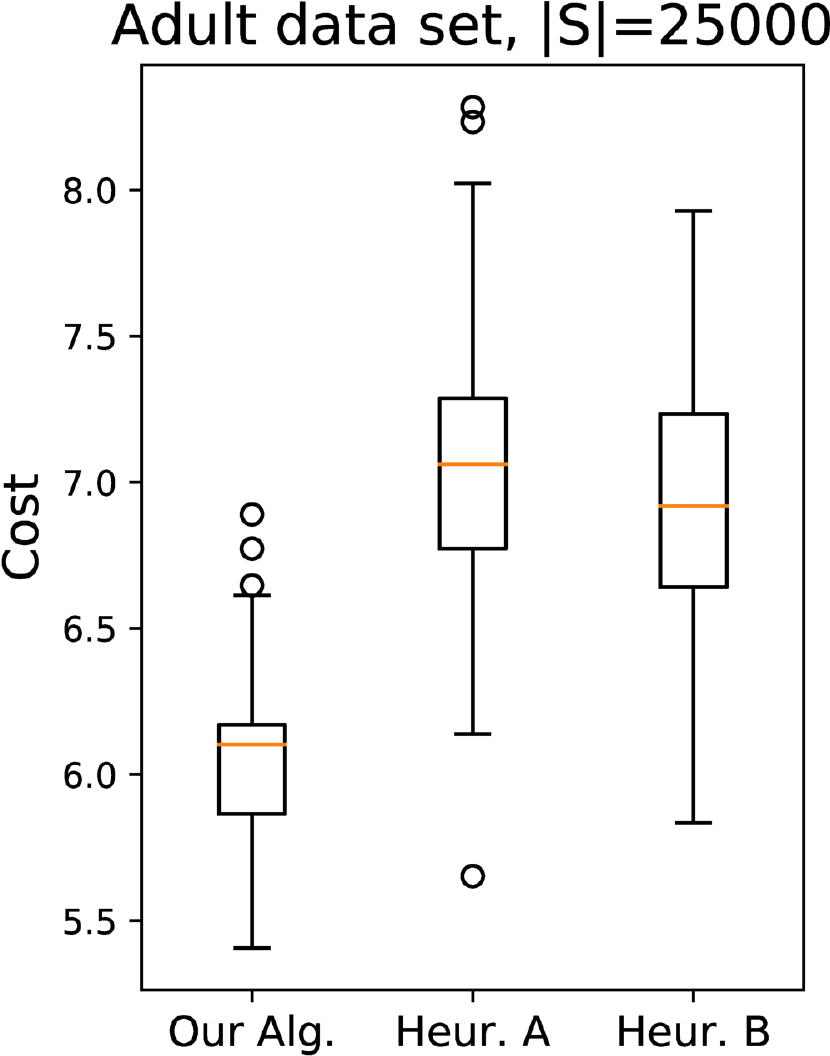}
\caption{
Similar 
experiments on the Adult data set as shown in Figure~\ref{fig_comp_heuristics}, but with different values of $k_{S_i}$. \textbf{1st plot:} $m=2$, $k_{S_1}=300$, $k_{S_2}=100$ ($S_1$ corresponds to male and $S_2$ to female). \textbf{2nd plot:}  $m=2$, $k_{S_1}=k_{S_2}=25$. \textbf{3rd plot:} $m=5$, $k_{S_1}=214$, $k_{S_2}=8$, $k_{S_3}=2$, $k_{S_4}=2$, $k_{S_5}=24$ ($S_1 \sim$ White, $S_2 \sim$ Asian-Pac-Islander, $S_3 \sim$ Amer-Indian-Eskimo, $S_4 \sim$ Other, $S_5 \sim$ Black). 
\textbf{4th plot:} $m=5$, $k_{S_1}=k_{S_2}=k_{S_3}=k_{S_4}=k_{S_5}=10$.
}\label{fig_comp_heuristics_SUPPMAT}
\end{figure}

\begin{figure*}[t]
\centering
\includegraphics[width=0.85\columnwidth]{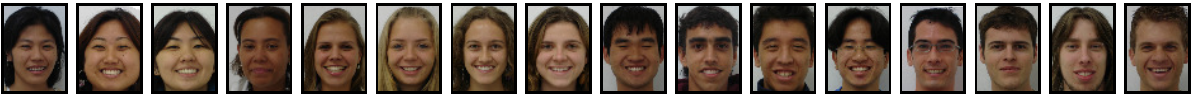}

\vspace{7mm}
\begin{tabular}{c|c|c}
Algorithm \ref{alg_greedy_standard} & Our Algorithm & \citet{celis2018}\\
\hline
~~~\includegraphics[width=0.19\columnwidth,trim={0 0 0 -0.2cm}]{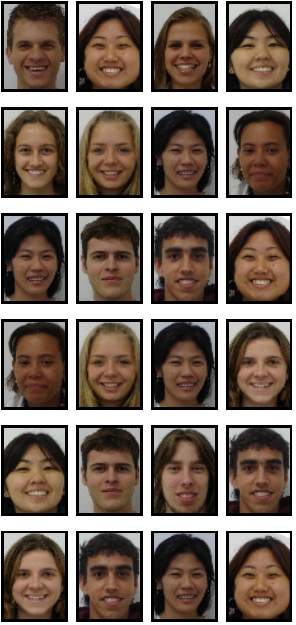}~~~~
&
~~~~\includegraphics[width=0.19\columnwidth]{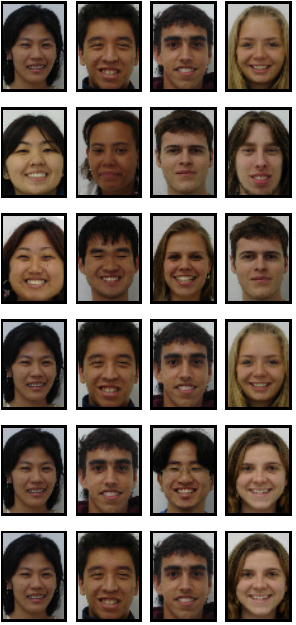}~~~~
&
~~~~\includegraphics[width=0.19\columnwidth]{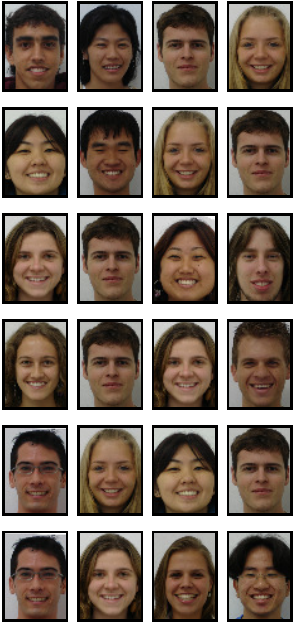}~~~
\end{tabular}
\caption{Similar experiment as shown in Figure~\ref{fig_exp_images}. A data set consisting of 16 images of faces (8 female, 8 male) and six summaries computed by the unfair Algorithm \ref{alg_greedy_standard}, our algorithm and the algorithm 
of 
\citet{celis2018}. The images are taken from the FEI face database 
available on \url{https://fei.edu.br/~cet/facedatabase.html}.
Note that in this experiment (and the one shown in Figure~\ref{fig_exp_images}) we are dealing with a  very small number of images  solely for the purpose of easy visual digestion. 
}\label{fig_exp_images_SUPPMAT}
\end{figure*}

\begin{figure}[h]
\centering
\includegraphics[scale=0.4]{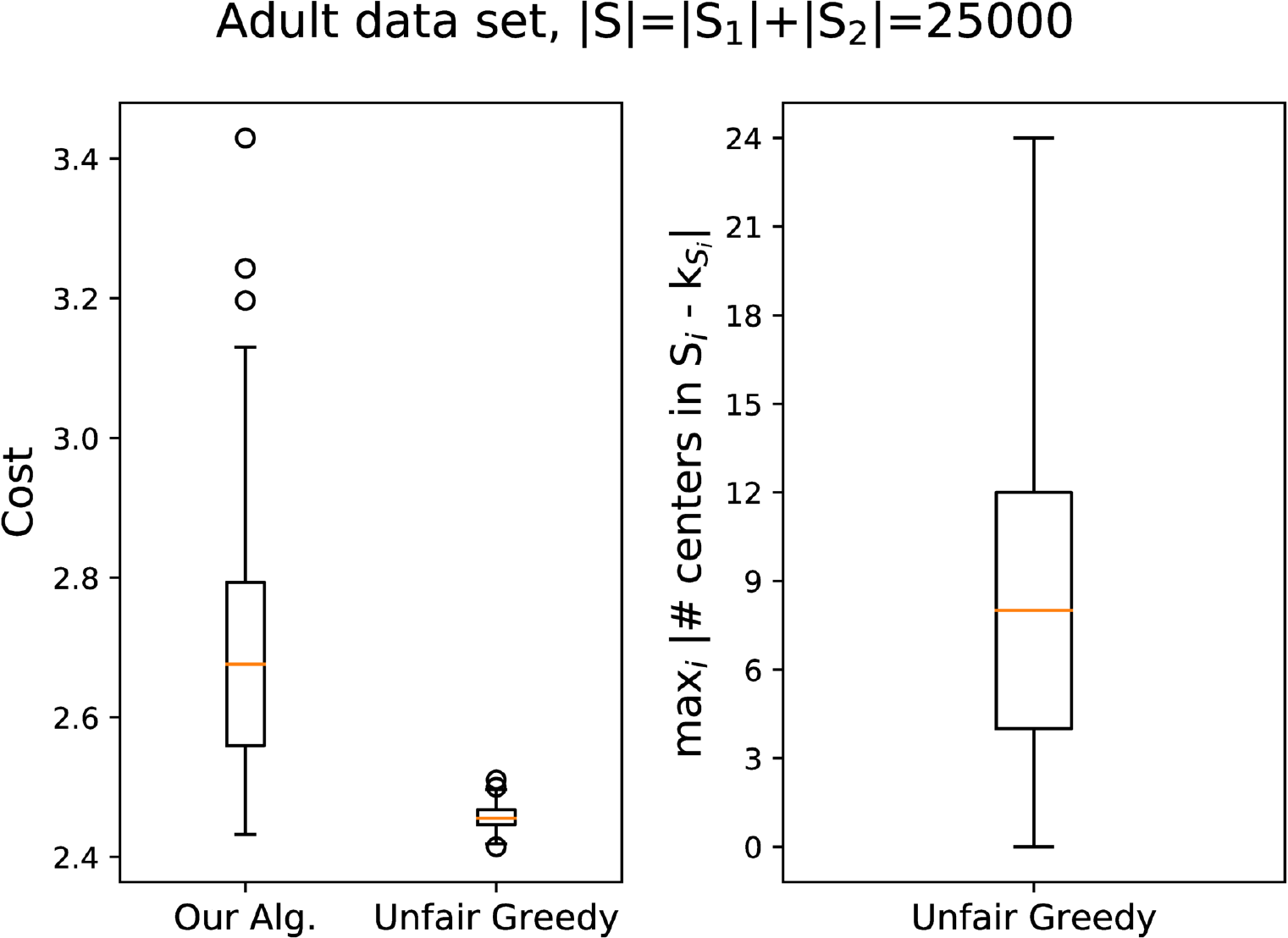}
\hspace{12mm}
\includegraphics[scale=0.4]{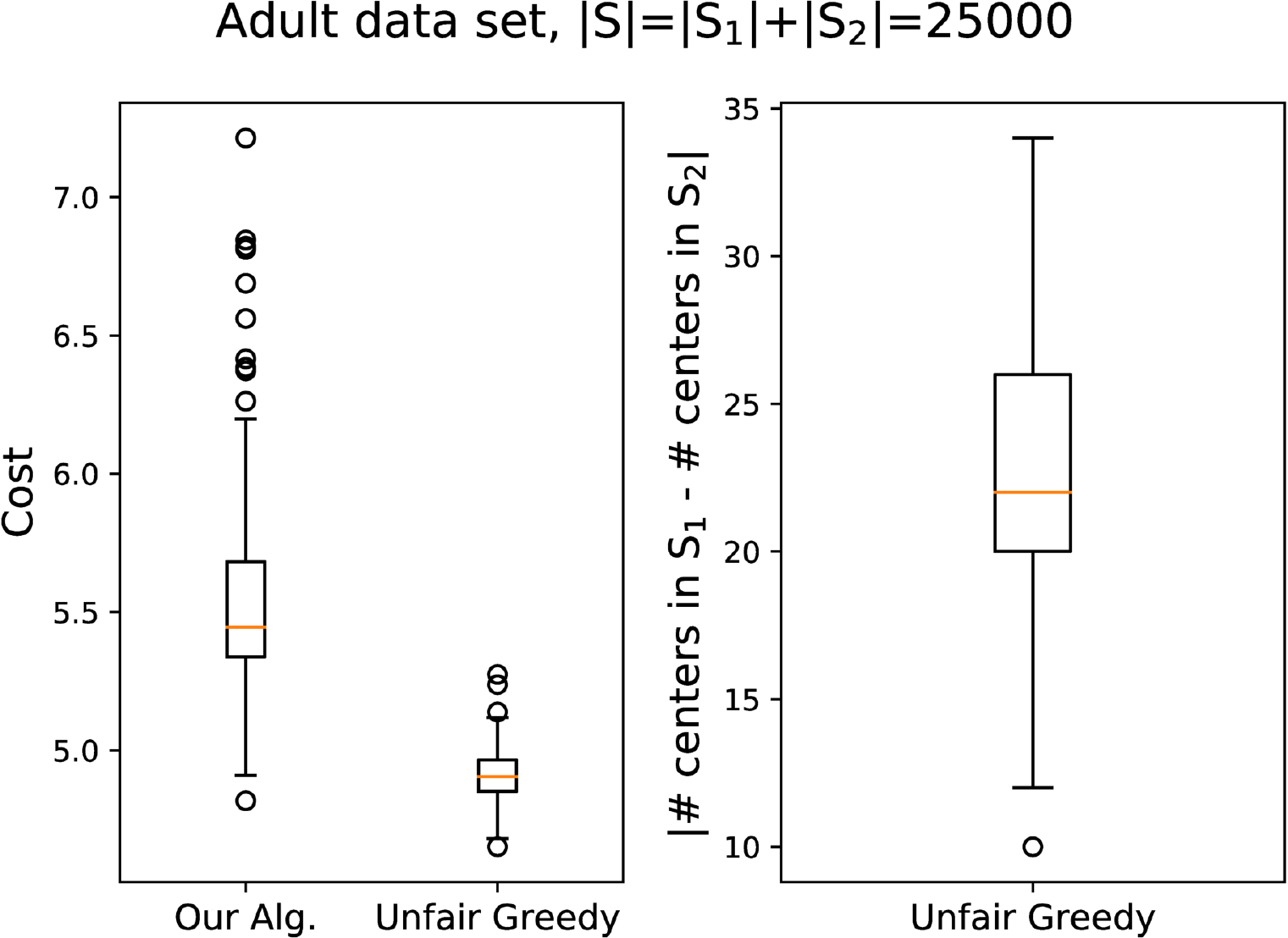}

\vspace{7mm}
\includegraphics[scale=0.4]{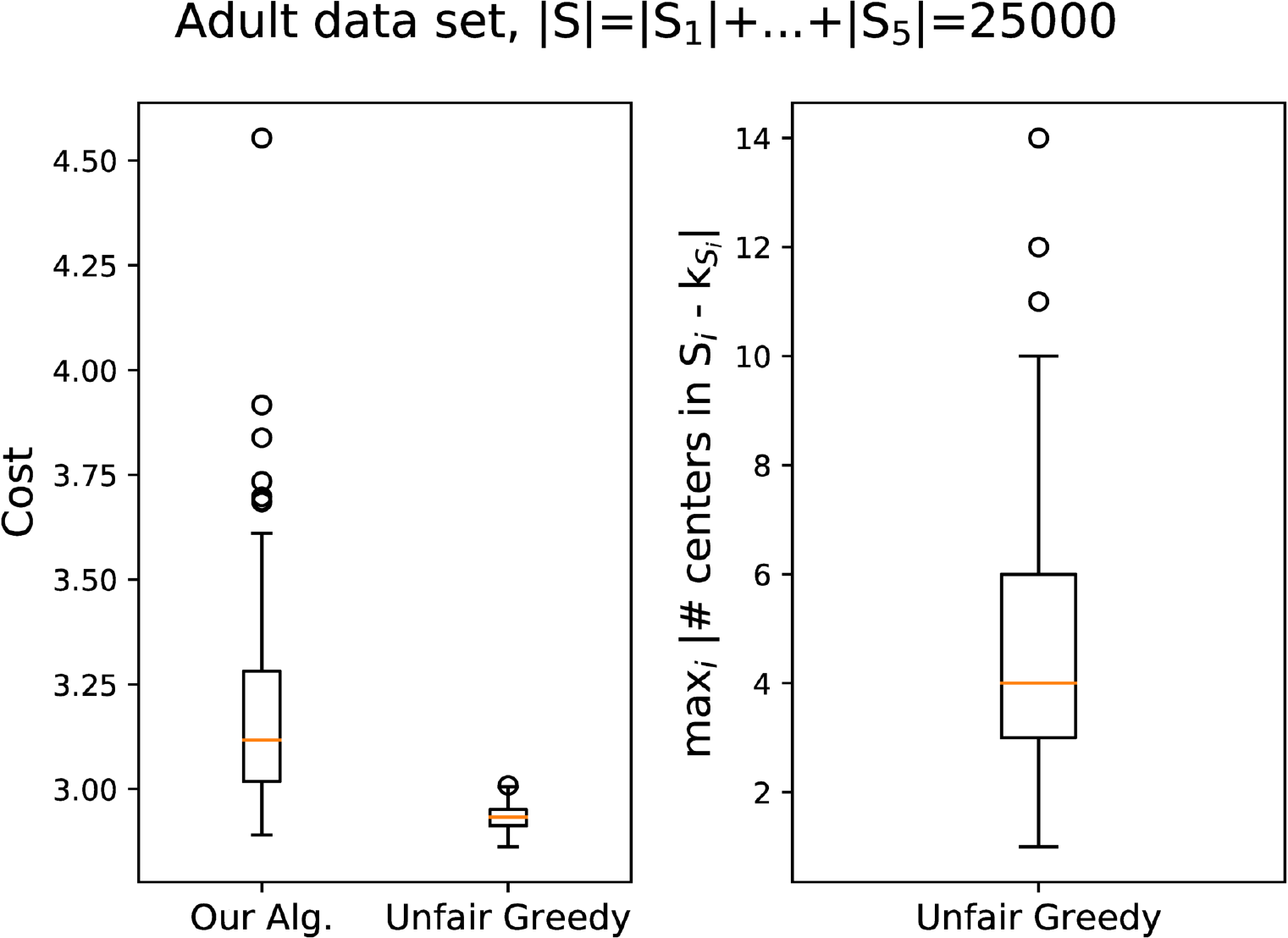}
\hspace{12mm}
\includegraphics[scale=0.4]{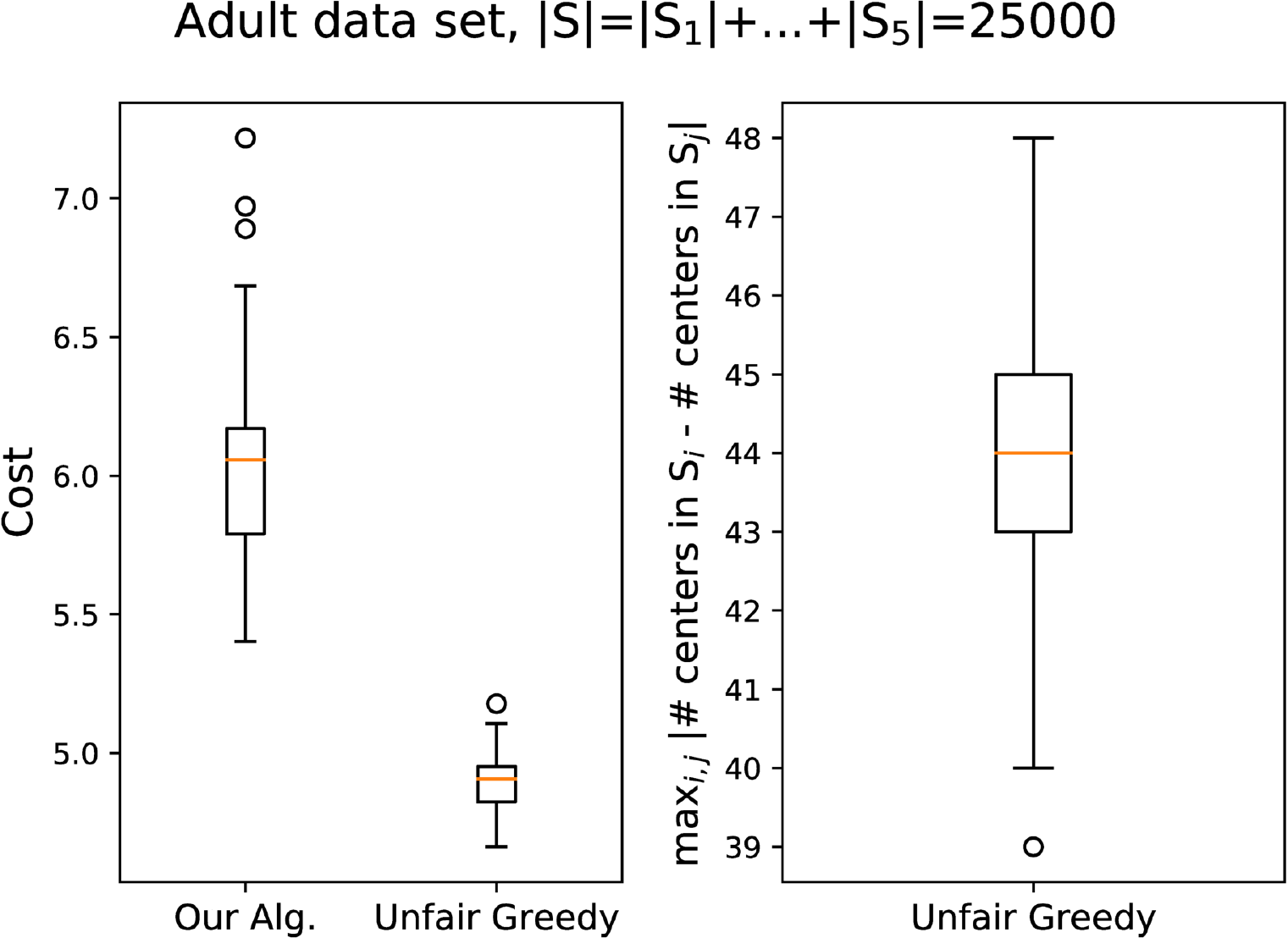}
\caption{
Similar 
experiments on the Adult data set as shown in Figure~\ref{fig_comp_greedy}, but with different values of $k_{S_i}$. \textbf{Top left:} $m=2$, $k_{S_1}=300$, $k_{S_2}=100$ ($S_1$ corresponds to male and $S_2$ to female). \textbf{Top right:}  $m=2$, $k_{S_1}=k_{S_2}=25$. \textbf{Bottom left:} $m=5$, $k_{S_1}=214$, $k_{S_2}=8$, $k_{S_3}=2$, $k_{S_4}=2$, $k_{S_5}=24$ ($S_1 \sim$ White, $S_2 \sim$ Asian-Pac-Islander, $S_3 \sim$ Amer-Indian-Eskimo, $S_4 \sim$ Other, $S_5 \sim$ Black). 
\textbf{Bottom right:} $m=5$, $k_{S_1}=k_{S_2}=k_{S_3}=k_{S_4}=k_{S_5}=10$.
}\label{fig_comp_greedy_SUPPMAT}
\end{figure}

In Figure~\ref{fig_exp_cost_comp_matr_intersection} we show the costs of the approximate solutions produced by our algorithm (Alg.~\ref{algorithm_mgroups}) and the algorithm by \citet{chen_matroid_center} (M.C.) in the run-time experiment shown in the right part of Figure~\ref{fig_exp_comparison_matroid_k_center}. 
In  Figure~\ref{fig_comp_heuristics_SUPPMAT}, Figure~\ref{fig_exp_images_SUPPMAT} and Figure~\ref{fig_comp_greedy_SUPPMAT} we provide similar experiments as shown in Figure~\ref{fig_comp_heuristics}, Figure~\ref{fig_exp_images}  and Figure~\ref{fig_comp_greedy}, respectively.

\vspace{8mm}

\vspace{1.4cm}

\vspace{4mm}


\end{document}